\documentclass{article}

\PassOptionsToPackage{numbers, compress}{natbib}
\PassOptionsToPackage{table, dvipsnames}{xcolor}

\usepackage[main, final]{neurips_2026}

\usepackage[utf8]{inputenc} 
\usepackage[T1]{fontenc}    
\usepackage{hyperref}       
\usepackage{url}            
\usepackage{booktabs}       
\usepackage{amsfonts}       
\usepackage{nicefrac}       
\usepackage{microtype}      
\usepackage{xcolor}         
\usepackage{multirow}
\usepackage[normalem]{ulem}
\usepackage{graphicx}
\usepackage{amsmath}
\usepackage{enumitem}

\title{OmniGF: A Dual-Branch Vision-Language Framework for Unified Gaze Following}

\author{%
  Qiaomu Miao\textsuperscript{1} \quad
  Haoyu Wu\textsuperscript{1} \quad
  Jingyi Xu\textsuperscript{1} \quad
  Minh Hoai\textsuperscript{2} \quad 
  Dimitris Samaras\textsuperscript{1} \\
  \textsuperscript{1}Stony Brook University \quad
\textsuperscript{2}	Adelaide University \quad \\
}

\begin{document}

\maketitle
\begin{abstract}
    Understanding human gaze behavior is essential for complex scene comprehension and human-computer interaction. Traditional gaze following models are typically restricted to pure spatial localization, lacking the high-level capacity to reason about semantic targets or complex social contexts. Furthermore, these models often process individuals sequentially, requiring redundant computations over the same scene image for multi-person inference. While recent Vision-Language Models (VLMs) offer the exceptional semantic reasoning needed to address gaze-related semantic tasks, their reliance on discrete text generation inherently limits precision in continuous spatial tasks like gaze localization. To bridge this gap, we propose OmniGF, a unified vision-language framework that adapts foundational VLMs for scalable multi-person gaze reasoning. The model adopts a dual-branch decoding strategy: a structured language branch generates discrete reasoning states, while a continuous spatial branch directly taps into the VLM's dense hidden states. Supervising these extracted representations with high-resolution gaze target heatmaps effectively overcomes the spatial bottleneck of text-only coordinate generation. Furthermore, to explicitly ground the model in multi-person scenes, we augment the input with head embeddings encoded from cropped head images, providing fine-grained appearance and orientation cues for all individuals simultaneously. By modeling all individuals and leveraging the strong semantic capability of VLMs, OmniGF seamlessly integrates precise spatial gaze target estimation, semantic gaze prediction, and complex social gaze reasoning. Extensive experiments demonstrate that our framework establishes new state-of-the-art performance across multiple standard benchmarks. Code is available at \url{https://github.com/cvlab-stonybrook/omnigf}.
\end{abstract}    
\section{Introduction}
\label{sec:intro}

Gaze following, which aims to infer where a person is looking within a scene, is a fundamental component of human social cognition and a critical task in computer vision. Accurately estimating human gaze targets is essential for a wide array of downstream applications, including human-robot interaction \cite{sheikhi2015combining, admoni2017social,saran2018human}, autonomous driving \cite{kasahara2022look, martin2018dynamics, liu2019gaze}, social behavioral analysis \cite{emery2000eyes, masse2017tracking}, and the diagnosis of neurodevelopmental disorders \cite{senju2009atypical, de2020computer}. Historically, the field has been dominated by specialized Convolutional Neural Networks (CNNs) \cite{NIPS2015_ec895663, chong2020detecting, fang2021dual, gupta2022modular} and Vision Transformers (ViTs) \cite{tafasca2024sharingan, gazelle}. While effective for basic spatial localization, these traditional models often process individuals sequentially, requiring redundant processing of the same scene image to predict gaze targets for multiple individuals. Furthermore, most methods are limited to pure spatial localization, lacking the high-level capacity to reason about the target semantics or to understand the broader context of human social interactions.

Recently, large Vision-Language Models (VLMs) have demonstrated unprecedented capabilities in generalized visual reasoning, achieving remarkable success across a variety of zero-shot and downstream tasks related to semantic reasoning, such as referring expression comprehension \cite{chen2023shikra} and semantic segmentation~\cite{beyer2024paligemma, glip}. However, trivially applying off-the-shelf VLMs to gaze following yields clearly suboptimal performance. Standard VLMs represent spatial coordinates using discrete language tokens, a paradigm that is inherently misaligned with the continuous, high-resolution precision required for dense spatial tasks like gaze localization. A concurrent work, Eyes on VLM~\cite{wang2026eyesonvlm}, shows that VLMs predicting gaze targets as text coordinates still lag behind specialized vision-based gaze models, even after task-specific fine-tuning. While a few pioneering works have attempted to bridge this gap, they remain limited. For instance, GazeVLM \cite{mathew2025gazevlm} relies on auxiliary depth map inputs yet fails to achieve competitive spatial precision against state-of-the-art vision-only models. Similarly, VL4Gaze \cite{wang2025vl4gaze} predominantly focuses on Gaze-oriented Visual Question Answering (VQA) and does not comprehensively address or evaluate standard 2D spatial gaze following benchmarks. Consequently, effectively harnessing the semantic power of VLMs for precise, multi-person gaze localization remains an open challenge.

To overcome these limitations, we propose OmniGF, a novel, unified Vision-Language framework for multi-person gaze target estimation and reasoning. Instead of forcing a VLM to regress spatial coordinates purely through discrete text tokens, we introduce a dual-branch decoding architecture. Our framework utilizes a fine-tuned structured language branch to generate discrete reasoning states (e.g., 2D target location, target in/out of frame, and semantic category), while simultaneously employing a continuous gaze branch dedicated to the dense spatial decoding of target heatmaps. This spatial decoder dynamically taps into the VLM's hidden state tokens derived from the generated gaze-related text for each person. By associating the visual embeddings of the scene with the gaze representations from the hidden states for each person, the model predicts gaze target heatmaps that directly benefit from the dense supervision signals, significantly improving spatial localization performance.

Furthermore, to ensure more efficient processing of complex scenes, OmniGF models multiple individuals jointly. To explicitly ground the VLM to each specific individual, we introduce a head-conditioned token injection mechanism. By embedding cropped visual head features directly into dedicated prompt token slots, we provide the backbone with the fine-grained appearance and orientation cues necessary to accurately disambiguate targets in dense environments. Crucially, by leveraging the strong semantic reasoning capacity of the VLM backbone, our framework seamlessly extends beyond simple coordinate regression. OmniGF jointly supports auxiliary tasks such as semantic gaze target recognition (identifying \textit{what} object is being looked at) and pairwise social gaze reasoning (e.g., shared attention and mutual gaze), unifying spatial, semantic, and social gaze understanding within a single model. 

In summary, our main contributions are as follows:

\textbullet\ \textbf{Unified Multi-Person VLM Framework.} We introduce OmniGF, a person-aware vision-language framework that models multiple individuals jointly via structured prompts, processing all persons together from a single VLM input.

\textbullet\ \textbf{Head-Conditioned Token Injection.} We propose a simple yet effective mechanism that injects head-centric visual embeddings into dedicated token slots, grounding the VLM with fine-grained appearance and orientation cues for each individual.

\textbullet\ \textbf{Dual-Branch Decoding for Reasoning and Spatial Prediction.} We design a dual-branch architecture that combines structured language generation for discrete reasoning with a gaze branch that extracts hidden states to predict dense, high-resolution gaze heatmaps.

\textbullet\ \textbf{Unified Gaze, Semantic, and Social Understanding.} Our framework jointly supports gaze target estimation, semantic gaze prediction, and pairwise social gaze reasoning within a single model, achieving state-of-the-art performance across five different benchmarks.

\section{Related Works}

\textbf{Gaze Following} aims to predict the location where a person is looking in a scene image. Early methods mostly adopted a two-branch design for gaze target prediction, with one branch encoding head pose and appearance and the other encoding scene saliency information~\cite{NIPS2015_ec895663, chong2018connecting, chong2020detecting, lian2018believe}. Later works leveraged additional modalities, such as depth and human pose, to further improve model performance~\cite{fang2021dual, Bao_2022_CVPR, gupta2022modular, miao2023patch, hu2022we, jin2022depth, tafasca2023childplay, yang2024aaai}. More recent models have incorporated large pretrained foundation models and achieved impressive performance on gaze following. Sharingan~\cite{tafasca2024sharingan} uses Multi-MAE~\cite{multimae} for multi-person gaze following, while Gaze-LLE~\cite{gazelle} employs DINOv2~\cite{oquab2023dinov2} with a head-prompting strategy and achieves further improvements. In parallel, several works have addressed gaze following from different perspectives, such as jointly predicting subject heads and their corresponding gaze targets~\cite{tu2022end, tonini2023object, tu2023joint}, incorporating audio cues for gaze following~\cite{hou2024multi}, or estimating gaze targets from multiple camera views~\cite{miao2025multi}. Meanwhile, recent works have explored jointly solving additional tasks together with gaze target estimation. Tafasca et al.~\cite{tafasca2024toward} explored joint prediction of the location and semantic label of the gaze target, while Gupta et al.~\cite{gupta2024mtgs} studied joint gaze target estimation and social gaze prediction. In this paper, we explore the use of multimodal LLMs for gaze target estimation and show that they can be seamlessly extended to semantic and social gaze prediction tasks as well.

\textbf{Vision-Language Models} (VLMs) like LLaVA~\cite{liu2023visual} and the Qwen-VL series~\cite{bai2023qwen, wang2024qwen2, bai2025qwen3} excel at generalized visual reasoning and have been successfully adapted for dense downstream tasks like visual grounding~\cite{rasheed2024glamm} and segmentation~\cite{beyer2024paligemma, glip, lai2024lisa}. However, their direct application to gaze following remains limited. Prior works primarily utilize VLMs for auxiliary purposes, such as extracting zero-shot contextual embeddings used for existing vision-based models~\cite{gupta2024exploring} or generating pseudo-annotations~\cite{miao2024diffusion}, rather than for direct localization. Concurrent VLM-based gaze models also exhibit notable limitations: GazeVLM~\cite{mathew2025gazevlm} relies on auxiliary depth inputs and lags behind specialized spatial models in precision, while VL4Gaze~\cite{wang2025vl4gaze} formulates the problem purely as VQA rather than evaluating on standard spatial localization benchmarks. Beyond generative VLMs, contrastive vision-language models such as CLIP~\cite{radford2021learning} have also been leveraged for gaze-related tasks: CLIP-Gaze~\cite{yin2024clipgaze} uses language-described gaze-irrelevant factors to learn domain-generalizable features for gaze direction estimation, and VL-HTR~\cite{wang2026vlhtr} introduces language-guided query alignment and gaze direction priors from pretrained VLMs for gaze target estimation and gaze object prediction. In contrast, our work directly adapts VLMs for standard gaze target estimation, demonstrating state-of-the-art effectiveness across continuous spatial localization, semantic prediction, and complex social gaze reasoning.

\section{Method}

Given an input image $I \in \mathbb{R}^{H \times W \times 3}$ and a set of $N$ head bounding boxes $\mathcal{B}=\{b_i\}_{i=1}^{N}$, where~$b_i \in \mathbb{R}^{4}$ denotes the head region of person $P_i$, our primary objective is to predict each person's gaze status (whether the target is inside or outside the frame) and to estimate the precise target coordinates if it falls within the image (\textit{where}). Furthermore, our framework seamlessly extends to higher-level relational tasks. For Semantic Gaze Reasoning, we infer the categorical identity of the object being attended to by each individual (\textit{what}). For Social Gaze Interaction, we analyze every pair of individuals $(P_i, P_j)$ to classify their relational dynamics into three distinct categories: Looking At Humans (LAH), Looking At Each Other (LAEO), and Shared Attention (SA).


\subsection{Overview}
\label{sec:method_overview}

\begin{figure}[t]
\centering
\includegraphics[width=\linewidth]{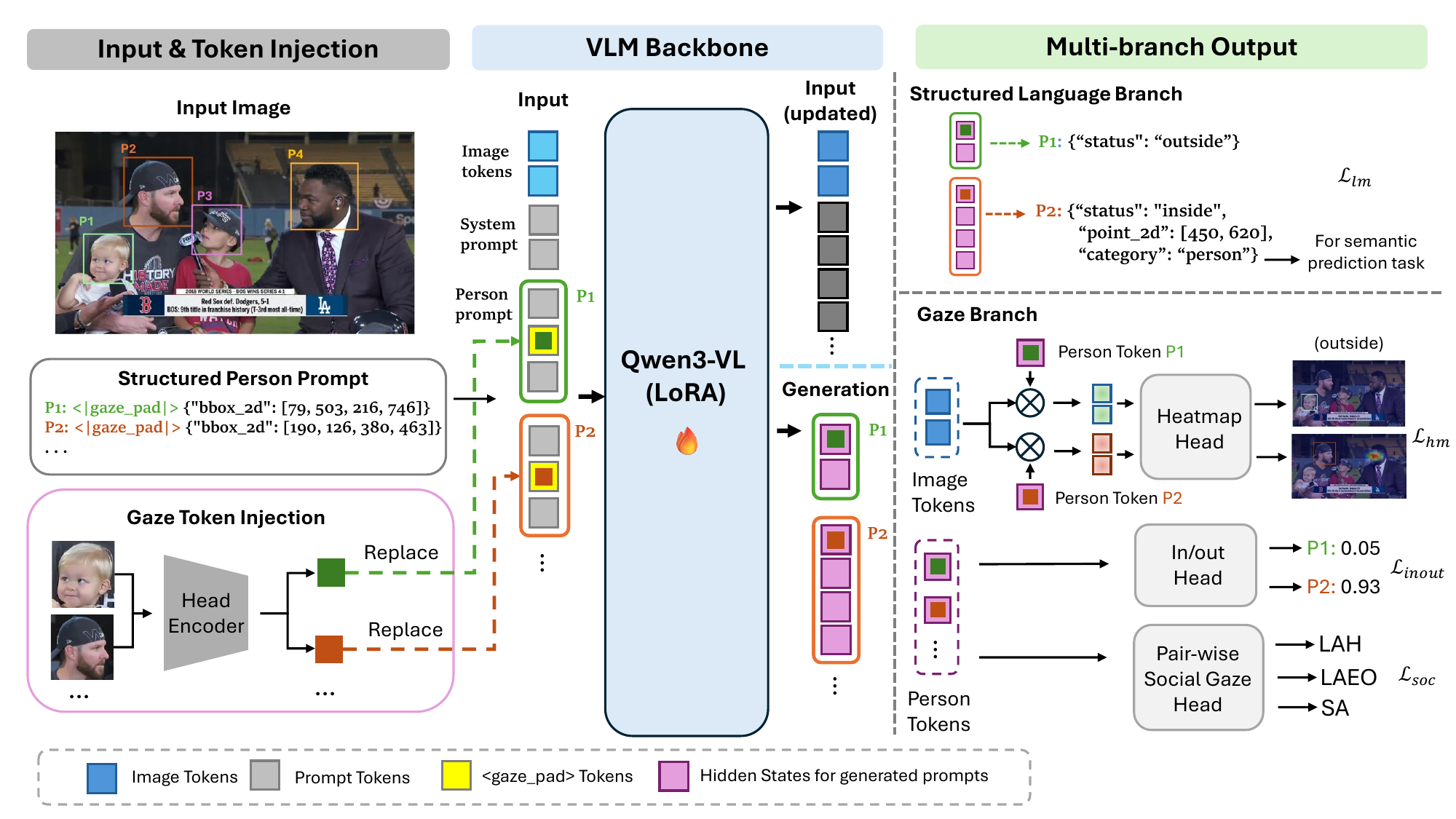}
\caption{
{\bf Overall framework}. Given an image with multiple head bounding boxes, we construct a structured person prompt that assigns each person a \texttt{<|gaze\_pad|>} placeholder followed by their head bounding box coordinates. The image, system prompt, and person prompt are processed by a LoRA-tuned Qwen3-VL. During token encoding, each \texttt{<|gaze\_pad|>} embedding is replaced with the corresponding head embedding extracted from the cropped head, thereby injecting person-specific visual cues into the VLM. The structured language branch generates JSON outputs for gaze status, target coordinates, and optional semantic categories. The gaze branch extracts image tokens and person tokens from the generated hidden states, where each person token is anchored to the first token of its corresponding generated output sequence. The image and person tokens are combined to predict per-person gaze heatmaps, while the person tokens are further used for in/out prediction and pair-wise social gaze reasoning. For clarity, only two example persons are shown.
}
\label{fig:framework}
\end{figure}

Our framework is built upon a LoRA~\cite{hu2022lora}-tuned Qwen3-VL~\cite{bai2025qwen3} backbone that processes the image, system instruction prompt (see Appendix~\ref{sec:appendix_prompts}), and structured multi-person prompts as input, where visual head embeddings are dynamically injected into the token sequence to ground each subject (Sec. \ref{sec:input_gaze_injection}). The architecture features a Structured Language Branch (Sec. \ref{sec:language_branch}), which leverages the VLM's discrete reasoning capacity to generate per-person textual predictions $\hat{T}_i$ including gaze status, coordinates, and semantic categories, and a Gaze Branch (Sec. \ref{sec:gaze_branch}), which decodes dense spatial signals from the backbone's hidden states to output high-resolution gaze heatmaps $M_i \in \mathbb{R}^{H' \times W'}$, along with binary in/out logits $p^{in}_i$, and pairwise social gaze logits $S_{i,j} \in \mathbb{R}^3$ from different prediction heads. By tapping into subject-specific hidden states and supervised with dense spatial heatmaps, this branch provides notably better localization than text-based coordinate generation. Therefore, we utilize the target derived from the dense heatmap $M_i$ as our final spatial gaze prediction. Fig.~\ref{fig:framework} presents an overview of our framework.

\subsection{Multimodal Input Construction and Gaze Token Injection}
\label{sec:input_gaze_injection}

Given an input image $I$ and a set of $N$ head bounding boxes $\mathcal{B}=\{b_i\}_{i=1}^{N}$, where $b_i=(x_i^1,y_i^1,x_i^2,y_i^2)$, we first construct a structured text prompt by defining a dedicated slot $T_i$ for each person using JSON format:  $\texttt{P}_i\texttt{:} \{\texttt{"bbox\_2d"}: [x_i^1,y_i^1,x_i^2,y_i^2]\}$.
In a standard VLM pipeline, the image $I$ is patchified and encoded by a vision encoder into a sequence of global vision tokens $H_{vis} \in \mathbb{R}^{V \times D}$, where $V$ is the number of image patches and $D$ is the hidden dimension. Simultaneously, the system instructions and the person prompts $\mathcal{T}=\{T_i\}_{i=1}^{N}$ are tokenized and embedded into text tokens $H_{text}$. These modalities are concatenated to form the initial sequence of hidden states $H^{(0)} \in \mathbb{R}^{L \times D}$ (where $L$ is the total sequence length) before being fed into the VLM transformer blocks. 

However, relying solely on this standard formulation presents two challenges for gaze target estimation. First, it forces the VLM to perform complex spatial referencing entirely through discrete text tokens. Second, the global vision tokens $H_{vis}$ may lack the fine-grained, face-centric appearance and 3D orientation cues necessary to disambiguate gaze directions in complex scenes. To resolve this, we introduce a \textit{head-conditioned gaze token injection} mechanism. We augment the input text by prepending a special placeholder token, \texttt{<|gaze\_pad|>}, to each person's prompt entry $ T_i$:
\begin{equation}
   \texttt{P}_i\texttt{:} \texttt{<|gaze\_pad|>}\ \{\texttt{"bbox\_2d"}: [x_i^1,y_i^1,x_i^2,y_i^2]\}.
\end{equation}
For each specified box $b_i$, we crop the corresponding head region $c_i$ from the input image $I$ and process it through a head encoder $E_{\mathrm{head}}$ to extract a gaze-aware visual embedding $e_i \in \mathbb{R}^D$. In our implementation, $E_{\mathrm{head}}$ utilizes a ResNet-18 backbone~\cite{resnet} pretrained on the Gaze360 dataset~\cite{gaze360}. 

Let $g_i$ denote the sequence index of the \texttt{<|gaze\_pad|>} token associated with person $P_i$. Prior to the first transformer layer of the LLM, we replace the text embedding at this index $H^{(0)}_{g_i}$ with our extracted head embedding $e_i$.
This targeted substitution explicitly injects localized head appearance and orientation cues into the sequence, grounding the person-specific prompt with rich visual features without disrupting the VLM's structured multimodal interface.

\subsection{Structured Language Branch}
\label{sec:language_branch}

The structured language branch utilizes the standard causal language modeling head of the VLM to autoregressively generate gaze predictions in a parseable JSON format. For each person $P_i$ specified in the multimodal prompt, the model predicts their gaze status (i.e., whether their gaze target falls inside or outside the image frame). If the target is inside the image, the model outputs the status alongside the normalized 2D target coordinates. For semantic gaze prediction tasks, the model optionally generates a text category describing the gazed-at object.

The generated text $\hat{T}_i$ for person $P_i$ follows a strict schema:
\begin{equation}
    \texttt{P}_i\texttt{:} \{\texttt{"status"}:\ \texttt{"inside"},\ \texttt{"point\_2d"}:\ [x, y],\ \texttt{"category"}:\ \texttt{"<object>"}\}
\end{equation}
Conversely, for targets located outside the image frame, the model simplifies the output schema to a single key-value pair: $\{\texttt{"status"}:\ \texttt{"outside"}\}$, omitting any further spatial or semantic attributes.

This text generation provides the discrete inference of the in/out status and the semantic category. Furthermore, the token sequence generated by this branch serves a critical structural purpose: the initial token of the output for each subject (e.g., the ``\texttt{P}'' token in ``\texttt{P}$_i$\texttt{:}'') acts as a spatial anchor. The hidden states at these specific positions provide person-specific embeddings rich in gaze-aware representations. As detailed in the following section, the gaze branch utilizes these specific token positions to extract person-aligned hidden states for dense geometric prediction. Here, the language branch primarily serves to enforce structured semantic reasoning and establish person-specific token anchors. We supervise the entire JSON text sequence using a standard autoregressive cross-entropy loss $\mathcal{L}_{lm}$. However, the 2D gaze target values generated within the text sequence are bypassed in favor of the gaze branch's dense predictions for the final localization result.

\subsection{Gaze Branch via Hidden State Extraction}
\label{sec:gaze_branch}

While the structured language branch outputs discrete text, continuous geometric and relational predictions are handled by the gaze branch using the model's internal representations. Specifically, we extract features from the hidden states of the final transformer layer in the VLM, denoted as $H^{(out)} \in \mathbb{R}^{L' \times D}$, where $L'$ corresponds to the total length of the input and generated tokens.

To perform person-specific spatial reasoning, we isolate the hidden state corresponding to the \texttt{P} token in the generated \texttt{P}$_i$\texttt{:} for each person as representative person embedding tokens. Because the VLM is instructed to generate personalized gaze information immediately following this token, the hidden state at this specific token position inherently aggregates and encodes rich, gaze-aware features for that individual. We denote this extracted person token as $h_{p_i} \in \mathbb{R}^{1 \times D}$. Simultaneously, we extract the sequence of updated image tokens $H'_{vis} \in \mathbb{R}^{V \times D}$ from $H^{(out)}$ to serve as the spatial canvas.

These extracted representations are then routed to specialized multi-task heads:

\textbullet\ \textbf{Gaze Target Estimation:} To predict the spatial target, the person token $h_{p_i}$ and the updated image tokens $H'_{vis}$ are fused via element-wise multiplication. This creates a personalized gaze feature for each person, which is then processed by a heatmap decoder consisting of several consecutive transposed convolutional layers to yield a high-resolution gaze target heatmap $M_i \in \mathbb{R}^{H' \times W'}$ for person $i$. This prediction is supervised by the Heatmap Loss $\mathcal{L}_{hm}$, which computes the Mean Squared Error (MSE) against a ground-truth heatmap generated by placing a 2D Gaussian distribution at the annotated target coordinate.

\textbullet\ \textbf{In/Out Prediction:} The person token $h_{p_i}$ is passed independently through a Multi-Layer Perceptron (MLP) to output a binary probability logit $p^{in}_i$, reinforcing the discrete inside/outside status generated by the language branch. This prediction is supervised by $\mathcal{L}_{inout}$, a Binary Cross-Entropy (BCE) loss computed against the ground-truth in/out label.

\textbullet\ \textbf{Social Gaze Reasoning:} For complex scene understanding, we evaluate pairwise relational interactions between individuals. For every pair of people $i$ and $j$, we concatenate their respective person tokens to form a joint representation $[h_{p_i}, h_{p_j}] \in \mathbb{R}^{1 \times 2D}$. This concatenated feature is passed through a three-layer MLP to estimate independent probability logits $S_{i,j} \in \mathbb{R}^3$ for three social gaze categories: LAH, LAEO, and SA. We apply a BCE loss independently to each logit to compute the overall Social Gaze Loss $\mathcal{L}_{soc}$.

\subsection{Training Objectives}
\label{sec:training_objectives}

The entire framework is trained end-to-end in a multi-task manner with LoRA fine-tuning. The overall training objective is defined as a weighted combination of the individual task losses:
\begin{equation}
\mathcal{L}_{total} = \lambda_{lm}\mathcal{L}_{lm} + \lambda_{hm}\mathcal{L}_{hm} + \lambda_{inout}\mathcal{L}_{inout} + \lambda_{soc}\mathcal{L}_{soc},    \end{equation}
where $\lambda_{lm}$, $\lambda_{hm}$, $\lambda_{inout}$, and $\lambda_{soc}$ are hyperparameters used to balance the multi-task contributions.

\section{Experiments}
\subsection{Datasets}
We conduct experiments on multiple benchmark datasets, which are described below.

\textbf{GazeFollow} \cite{NIPS2015_ec895663} is a large-scale image benchmark for gaze following. It comprises around 130K person instances across 122K images. The test set features 4,782 gaze instances, with each target labeled by roughly 10 independent annotators. Furthermore, we utilize the semantic extension introduced by Tafasca et al.~\cite{tafasca2024toward} for the semantic prediction task, which augments the dataset with target object categories. This extension provides pseudo-labels over about 3,500 categories for training and a manually curated test set spanning a 346-class vocabulary for semantic evaluation.

\textbf{VideoAttentionTarget}~\cite{chong2020detecting} is a video gaze following dataset consisting of 1,331 clips sourced from 50 YouTube TV shows. It provides 164K annotated instances across 71K frames, notably introducing binary inside/outside frame labels for the gaze targets. The test split was constructed by holding out 10 complete TV shows, yielding 31,978 test annotations.

\textbf{ChildPlay}~\cite{tafasca2023childplay} is a video dataset focusing on the gaze behaviors of children during social interactions. Sourced from 95 YouTube videos, it yields 401 short clips containing approximately 258K instances across 120K frames. Beyond standard head bounding boxes and 2D target coordinates, instances are annotated with gaze state classes, including target inside, target outside, and gaze shifts.

\textbf{GazeHOI}~\cite{tafasca2024toward} is a benchmark designed for simultaneous spatial localization and semantic classification. Rather than relying on automatically labeled annotations, GazeHOI was constructed by repurposing five established Human-Object Interaction (HOI) datasets, isolating specific interaction verbs where a person's gaze is directed at the interacted object. In total, it provides 55,995 gaze instances across 463 distinct object categories, and we evaluate on its test split of 4,934 (person, object) pairs.

\textbf{VSGaze}~\cite{gupta2024mtgs} is a composite dataset specifically designed for social gaze prediction, aggregating data from VideoAttentionTarget, ChildPlay, VideoCoAtt \cite{fan2018inferring}, and UCO-LAEO \cite{marin2019laeo}. The authors unified these disparate sources through a cross-annotation pipeline. First, they generated continuous head track annotations by detecting missing heads and tracking them across frames. Second, they cross-annotated the tasks: social gaze datasets (VideoCoAtt, UCO-LAEO) were augmented with spatial gaze targets approximated at bounding box centers, while the gaze following datasets were extended with social gaze labels. The resulting unified dataset contains 714K annotated gaze points with 444k, 30k, 420k positive pair-wise social gaze labels for LAH, LAEO, and SA respectively.

\subsection{Evaluation Metrics}

For gaze target localization, we compute the L2 distance (\textbf{Dist.}) on a normalized $1{\times}1$ image plane, extracting the predicted coordinate from the peak heatmap activation. To account for multiple annotators in GazeFollow, we report both \textbf{Avg Dist.}, the distance to the average of the annotated points, and \textbf{Min Dist.}, the distance to the closest annotated point. On GazeFollow, we additionally report the heatmap \textbf{AUC}. Following \cite{tafasca2024sharingan, cao2026gaze}, we do not report AUC on the single-annotation video benchmarks (VideoAttentionTarget and ChildPlay), where the ground-truth heatmap is an artificial Gaussian centered on a single annotated point and AUC is therefore uninformative \cite{tafasca2024sharingan}. AUC on all benchmarks is reported in Appendix~\ref{sec:appendix_auc_plah} for completeness. We additionally report the precision of looking at heads (\textbf{PLAH}) \cite{tafasca2024sharingan} on all benchmarks, where a prediction is counted as positive only if the predicted and annotated gaze points fall within the same head box. PLAH values of Gaze-LLE are computed by us using its official checkpoints and the evaluation protocol of Sharingan~\cite{tafasca2024sharingan}, while those of other methods are taken from~\cite{tafasca2024sharingan}. Target inside/outside prediction is evaluated via \textbf{Average Precision (AP)}.

For semantic gaze prediction on GazeFollow, we report top-1 accuracy (\textbf{Accuracy@1}) and \textbf{MultiAccuracy@1}, which accounts for multiple labels that can be assigned to the same subject. Because our framework generates text for semantic labels rather than probability logits, we omit Accuracy@3. On GazeHOI, we evaluate target localization using \textbf{GazeAcc}, which measures whether the predicted point falls within the ground-truth object bounding box, along with \textbf{Accuracy@1} for recognition.

For social gaze prediction on VSGaze, we follow the standard pairwise evaluation protocol established by MTGS \cite{gupta2024mtgs}, reporting \textbf{F1} scores for LAH and LAEO, and \textbf{AP} for SA.

\subsection{Implementation Details} \label{sec:impl_details}
Our framework utilizes the Qwen3-VL-4B-Instruct backbone, adapted via LoRA ($r=16, \alpha=32$) applied to the query and value projections of all 36 transformer layers of the language model. The head bounding box coordinates are normalized to the relative coordinate range $[0, 999]$ in the input prompt following the grounding convention of Qwen3-VL. Input resolutions are fixed at $448 \times 448$ for the global scene and $224 \times 224$ for cropped heads, and gaze heatmaps are predicted at $64 \times 64$ for each person. We use the AdamW optimizer with a batch size of 8 and two gradient accumulation steps (effective batch size 16). On GazeFollow, we train for 20 epochs with learning rates of $10^{-4}$ for the LoRA adapter and $2.5 \times 10^{-4}$ for the head encoder and prediction heads, using one epoch of linear warmup followed by cosine annealing. The loss coefficients are set to $\lambda_{lm} = 1.0$, $\lambda_{hm} = 10.0$, and $\lambda_{soc} = 1.0$ (when applicable), and $\lambda_{inout} = 0.1$ when training with in/out supervision. As the standard GazeFollow annotations contain only in-frame gaze targets, the GazeFollow model in Table~\ref{tab:gf_vat_results} is trained with $\lambda_{inout} = 0$. The models for VideoAttentionTarget, ChildPlay, VSGaze, and the semantic gaze prediction models are fine-tuned from the model pretrained on GazeFollow. The fine-tuning learning rates, schedules, and checkpoint selection for each task are detailed in Appendix~\ref{sec:appendix_impl}. Experiments are conducted on a single NVIDIA H100 (80 GB) or NVIDIA RTX PRO 6000 Blackwell (96 GB) GPU. Training times vary from 2 to 30 hours across datasets.

\begin{table}[t]
\centering
\small
\caption{Comparison with state-of-the-art methods on GazeFollow and VideoAttentionTarget. In the column of input modality, I is image, D is depth, P is pose, and E is eyes. Multi-person denotes support for multi-person inference in a single forward pass. AUC is reported only on GazeFollow, which has multiple annotations per instance; AUC on all benchmarks is reported in Appendix~\ref{sec:appendix_auc_plah}. Best numbers are marked as bold while second-best are underlined.}
\label{tab:gf_vat_results}
\setlength{\tabcolsep}{3pt}
\begin{tabular}{@{}llcccccccc@{}}
\toprule
\multirow{2}{*}{Method} & \multirow{2}{*}{Input} & \multirow{2}{*}{\shortstack{Multi-\\person}}
& \multicolumn{4}{c}{GazeFollow}
& \multicolumn{3}{c}{VideoAttentionTarget} \\
\cmidrule(lr){4-7} \cmidrule(lr){8-10}
& & & Avg Dist.$\downarrow$ & Min Dist.$\downarrow$ & AUC$\uparrow$ & PLAH$\uparrow$
& Dist.$\downarrow$ & AP$\uparrow$ & PLAH$\uparrow$ \\
\midrule
Recasens \cite{NIPS2015_ec895663}       & I       & $\times$   & 0.190 & 0.113 & 0.878 & --   & --    & --    & --    \\
Chong \cite{chong2018connecting}         & I       & $\times$   & 0.187 & 0.112 & 0.896 & --   & 0.171 & 0.712 & --    \\
Lian \cite{lian2018believe}          & I       & $\times$   & 0.145 & 0.081 & 0.906 & --    & --    & --    & --    \\
Chong \cite{chong2020detecting} & I       & $\times$   & 0.137 & 0.077 & 0.921 & --    & 0.134 & 0.853 & --    \\
Fang \cite{fang2021dual}          & I+D+E   & $\times$   & 0.124 & 0.067 & 0.922 & --    & 0.108 & 0.896 & --    \\
Jin  \cite{jin2021multi}          & I       & \checkmark & 0.126 & 0.076 & 0.919 & --    & 0.134 & 0.880 & --    \\
Bao  \cite{Bao_2022_CVPR}          & I+D+P   & $\times$   & 0.122 & --    & 0.928 & --    & 0.120 & 0.869 & --    \\
Jin  \cite{jin2022depth}           & I+D+P   & $\times$   & 0.118 & 0.063 & 0.920 & --    & 0.104 & 0.895 & --    \\
Gupta  \cite{gupta2022modular}         & I+D+P   & $\times$   & 0.114 & 0.056 & 0.943 & --    & 0.110 & 0.879 & --    \\
Miao  \cite{miao2023patch}          & I+D     & $\times$   & 0.123 & 0.065 & 0.934 & --    & 0.109 & \uline{0.908} & --    \\
Tafasca \cite{tafasca2023childplay}       & I+D     & $\times$   & 0.125 & 0.064 & 0.936 & 0.622 & 0.109 & 0.834 & \uline{0.752} \\
Sharingan \cite{tafasca2024sharingan}       & I       & \checkmark & 0.113 & 0.057 & 0.944 & 0.667 & 0.107 & 0.891 & 0.738 \\
MTGS \cite{gupta2024mtgs}                 & I       & \checkmark & 0.116 & 0.059 & 0.929 & --    & 0.105 & 0.869 & --    \\ 
Gaze-LLE (ViT-B) \cite{gazelle}    & I       & \checkmark & 0.104 & 0.045 & \uline{0.956} & 0.665 & 0.107 & 0.897 & 0.722 \\
Gaze-LLE (ViT-L) \cite{gazelle}     & I       & \checkmark & \uline{0.099} & \uline{0.041} & \textbf{0.958} & \uline{0.679} & 0.103 & 0.903 & 0.738 \\
GazeVLM \cite{mathew2025gazevlm}  & I+D & \checkmark  & 0.123 & 0.061 & 0.936 & --    & \uline{0.102} & 0.898 & --    \\
\midrule
OmniGF                  & I       & \checkmark & \textbf{0.091} & \textbf{0.040} & 0.941 & \textbf{0.753} & \textbf{0.096} & \textbf{0.923} & \textbf{0.759} \\
\midrule
Human                 & --      & --         & 0.096 & 0.040 & 0.924 & --    & 0.051 & 0.925 & --    \\
\bottomrule
\end{tabular}
\end{table}

\begin{table}[t]
\centering
\small
\begin{minipage}[t]{0.47\textwidth}
\centering
\caption{Comparison Results on ChildPlay.}
\label{tab:childplay_results}
\setlength{\tabcolsep}{3.5pt}
\begin{tabular}{lccc}
\toprule
Method & Dist.$\downarrow$ & AP$\uparrow$ & PLAH$\uparrow$ \\
\midrule
Gupta \cite{gupta2022modular} & 0.113 & 0.983 & -- \\
Tafasca  \cite{tafasca2023childplay} & 0.107 & 0.986 & 0.590 \\
Sharingan \cite{tafasca2024sharingan} & 0.106 & 0.990 & 0.600 \\
MTGS \cite{gupta2024mtgs} & 0.117 & 0.993 & -- \\
Gaze-LLE (ViT-B) \cite{gazelle} & 0.106 & \uline{0.994} & 0.579 \\
Gaze-LLE (ViT-L) \cite{gazelle} & \uline{0.101} & \uline{0.994} & \uline{0.612} \\
\midrule
OmniGF & \textbf{0.090} & \textbf{0.996} & \textbf{0.662} \\
\bottomrule
\end{tabular}
\end{minipage}
\hfill
\begin{minipage}[t]{0.47\textwidth}
\centering
\caption{Ablation Results on GazeFollow.}
\label{tab:ablation_components}
\setlength{\tabcolsep}{2.5pt}
\begin{tabular}{@{}cccccc@{}}
\toprule
\multicolumn{3}{c}{Components} & \multicolumn{3}{c}{Metrics} \\
\cmidrule(lr){1-3} \cmidrule(lr){4-6}
\begin{tabular}{@{}c@{}}Language\\(LoRA)\end{tabular} &
\begin{tabular}{@{}c@{}}Spatial\\Decoding\end{tabular} &
\begin{tabular}{@{}c@{}}Token\\Inject.\end{tabular} &
Avg $\downarrow$ & Min $\downarrow$ & AUC $\uparrow$ \\
\midrule
 & & & 0.340 & 0.258 & -- \\
\checkmark & & & 0.114 & 0.055 & -- \\
 & \checkmark & & 0.167 & 0.103 & 0.904 \\
\checkmark & \checkmark & & 0.104 & 0.047 & 0.939 \\
\checkmark & \checkmark & \checkmark & \textbf{0.091} & \textbf{0.040} & \textbf{0.941} \\
\bottomrule
\end{tabular}
\end{minipage}

\end{table}

\subsection{Comparison with State-of-the-Art}
\label{sec:sota_comparison}

\begin{figure}[t]
\setlength\tabcolsep{3pt}
\centering
\begin{tabular}{cccc}
 \includegraphics[width=0.24\linewidth, height=0.8in]{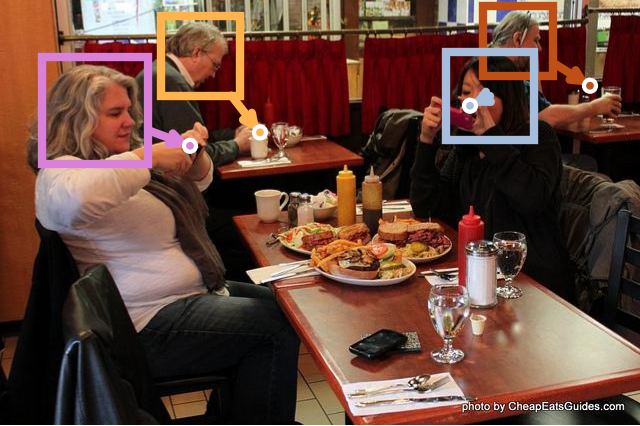} &
 \includegraphics[width=0.24\linewidth, height=0.8in]{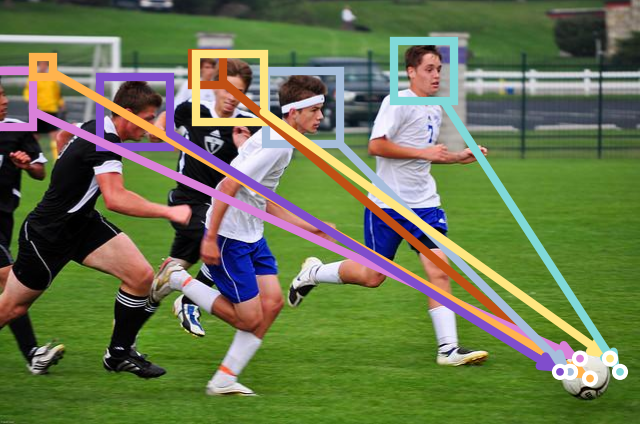} &
 \includegraphics[width=0.24\linewidth, height=0.8in]{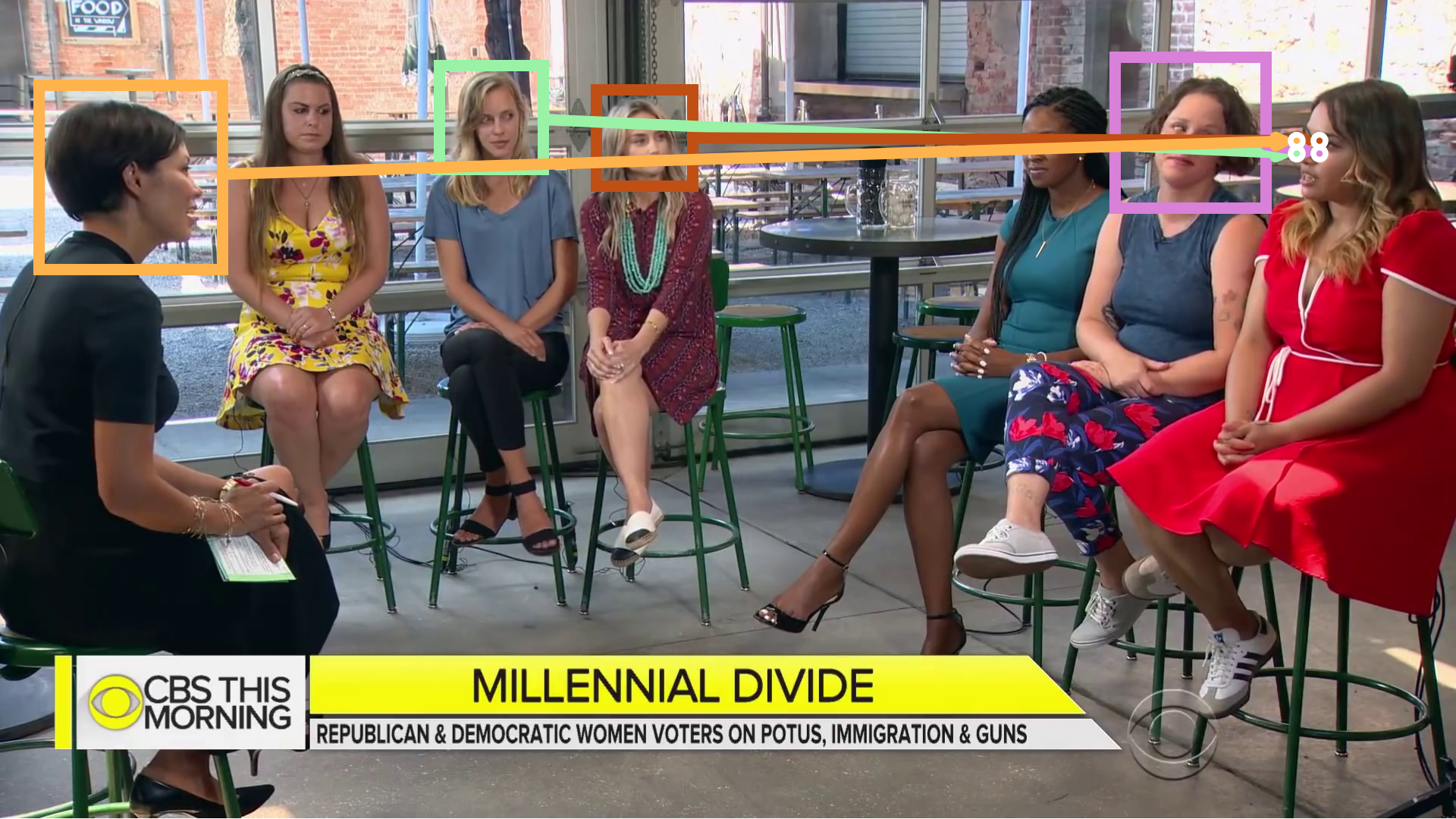} &
 \includegraphics[width=0.24\linewidth, height=0.8in]{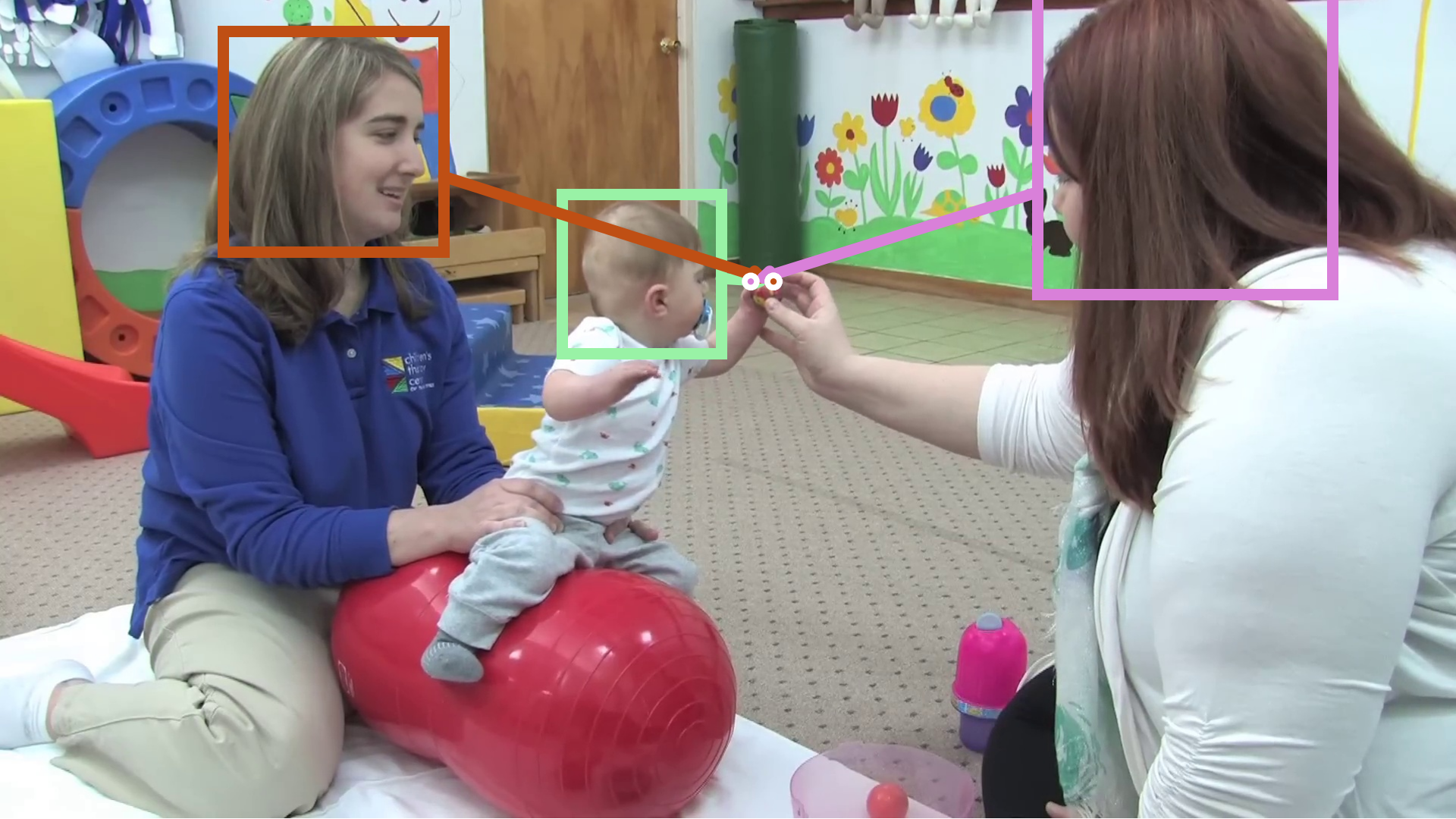} \\
 \includegraphics[width=0.24\linewidth, height=0.8in]{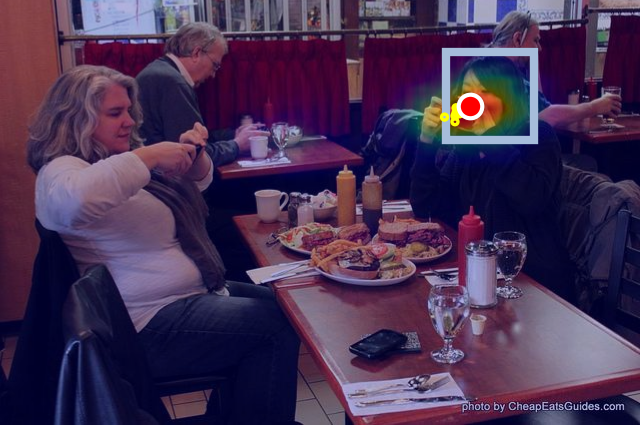} &
 \includegraphics[width=0.24\linewidth, height=0.8in]{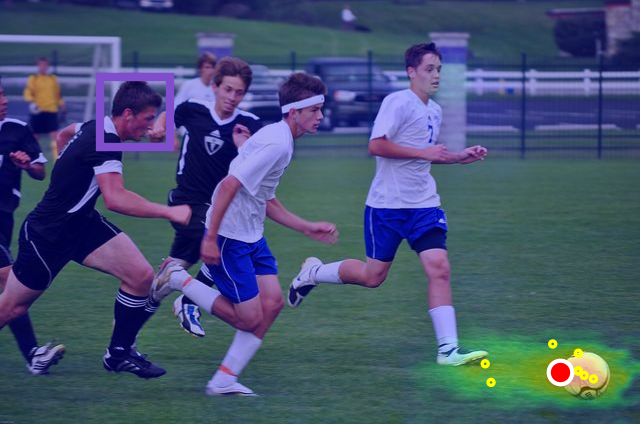} &
 \includegraphics[width=0.24\linewidth, height=0.8in]{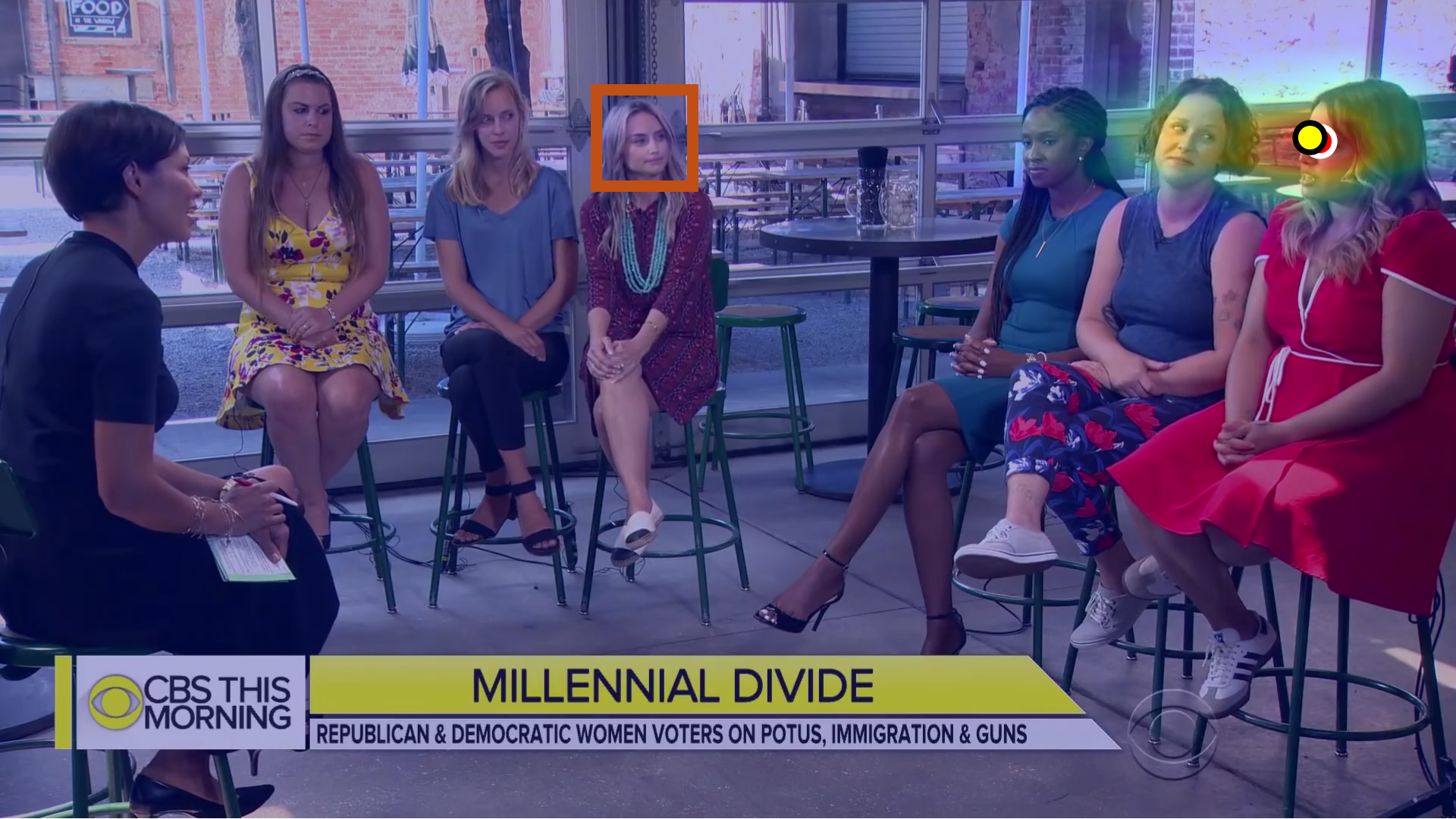} &
 \includegraphics[width=0.24\linewidth, height=0.8in]{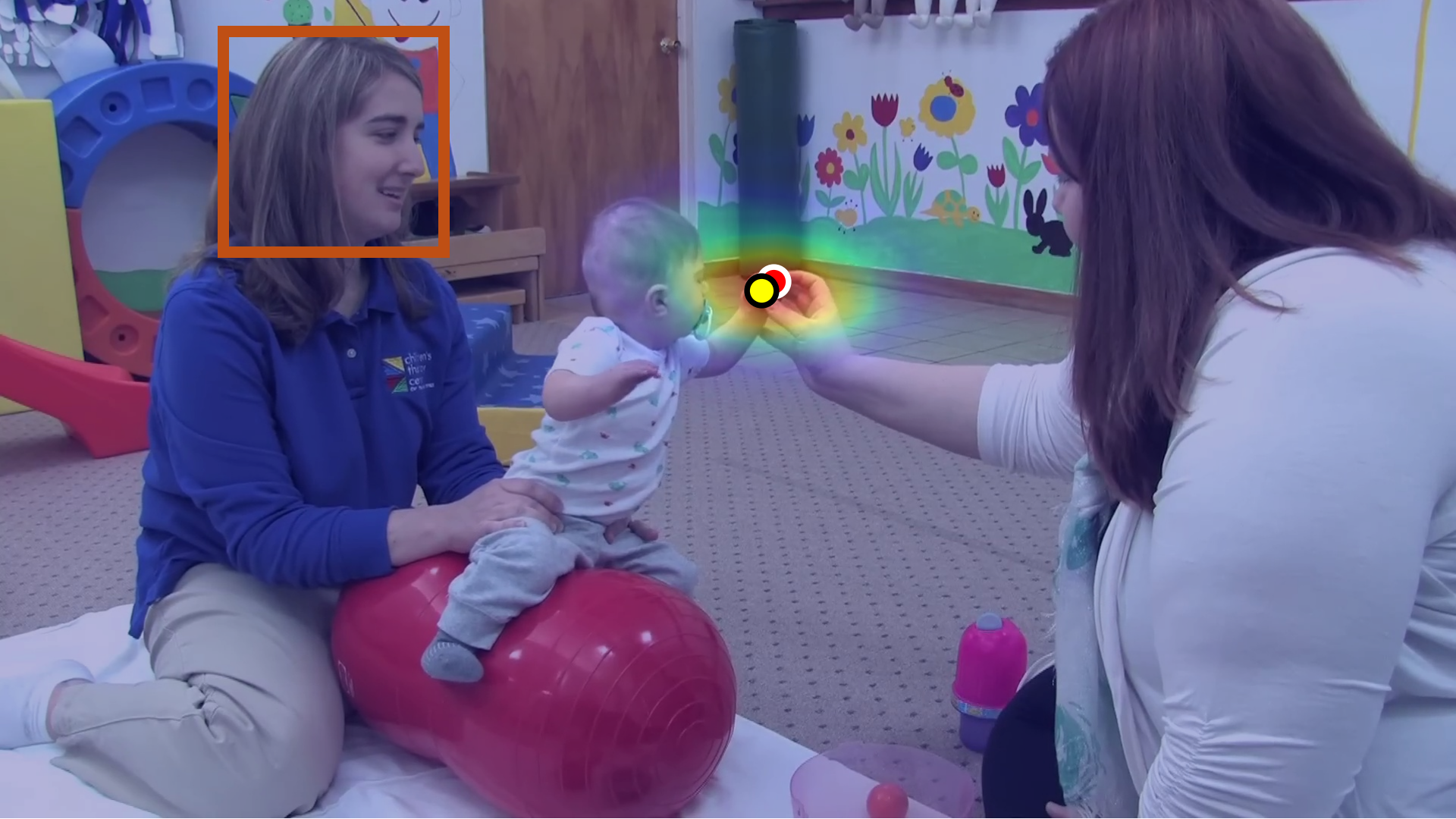} \\
\end{tabular}
\caption{Visualizations of model predictions of OmniGF on GazeFollow (1st and 2nd columns), VideoAttentionTarget (3rd column) and ChildPlay (4th column). The first row shows predicted gaze target locations for multiple persons in an image. The 2nd row shows the predicted heatmap for a representative person in the image. Ground truth annotations are visualized in yellow, and the predicted point is visualized in red.}
\label{fig:qualitative_vis}
\end{figure}

We compare our proposed framework against existing state-of-the-art gaze following methods on the GazeFollow and VideoAttentionTarget datasets in Table~\ref{tab:gf_vat_results}, and on the ChildPlay dataset in Table~\ref{tab:childplay_results}. Our framework establishes a new state-of-the-art across all distance-based target estimation and in/out prediction metrics on all datasets. On GazeFollow, our method achieves an Avg Dist. of 0.091 and a Min Dist. of 0.040, surpassing the reported human performance baseline for Avg Dist. (0.096), while remaining competitive in AUC (0.941). OmniGF also achieves the highest PLAH on all three datasets. We note that AUC has been shown to favor heatmaps with multiple modes over precise localization of the gaze target \cite{miao2024diffusion, tonini2023object}. We therefore regard the distance metrics, which directly measure geometric localization error, as the primary indicator of performance. On VideoAttentionTarget, our model achieves the lowest spatial error (Dist. 0.096) and highly competitive in/out prediction performance (AP 0.923), closing the gap toward human-level recognition. Crucially, our model achieves these results by processing all people with a single VLM input using solely RGB image inputs. This significantly outperforms complex architectures that rely on auxiliary modalities (e.g., depth, pose) and establishes a clear performance margin over prior multi-person capable models. Furthermore, our model significantly outperforms GazeVLM, a concurrent VLM-based baseline that requires depth images as an additional input modality. Meanwhile, we validate our model's robustness on the ChildPlay dataset, where it achieves a new state-of-the-art Dist. of 0.090 and an AP of 0.996, outperforming recent frameworks like Sharingan and Gaze-LLE. Qualitative results in Fig.~\ref{fig:qualitative_vis} show high-quality multi-person gaze predictions from our model across multiple datasets.

\subsection{Ablation Study}
\label{sec:ablation}

To validate our architectural innovations, we conduct a comprehensive ablation study on the GazeFollow dataset (Table~\ref{tab:ablation_components}). We first evaluate the pretrained Qwen3-VL \cite{bai2025qwen3} backbone in a zero-shot setting. Using our exact system and person prompts to instruct the model to generate gaze status and target coordinates in text, it yields very poor localization (0.340 Avg Dist.). This confirms that off-the-shelf text generation struggles with precise spatial localization for gaze following. Fine-tuning the structured language branch via LoRA significantly reduces this error (0.114) but remains limited by the precision bottleneck of text-based target localization. Conversely, training only the continuous gaze branch for dense heatmap decoding proves suboptimal (0.167). Because the heatmap head relies on person token embeddings extracted from the prompt, failing to simultaneously fine-tune the language branch deprives these embeddings of the necessary gaze-related context for accurate heatmap prediction. However, combining both approaches into our dual-branch architecture overcomes these limitations (0.104), proving that the hidden state tokens produced during language generation serve as essential structural anchors for extracting high-resolution spatial heatmaps. Finally, integrating dynamic gaze token injection achieves our state-of-the-art performance (0.091 Avg Dist.). By injecting cropped visual head embeddings directly into the input sequence, this mechanism provides the explicit, fine-grained appearance and orientation cues necessary to accurately disambiguate targets. Further ablations, zero-shot results of other VLMs, a runtime analysis, and qualitative results of semantic and social gaze prediction are provided in Appendices~\ref{sec:appendix_person_order}--\ref{sec:qualitative_supp}.


\subsection{Cross-dataset Evaluation}
\label{sec:cross_dataset}

To assess generalization to unseen data, we follow the zero-shot protocol of prior work \cite{gazelle}: we train OmniGF on GazeFollow and evaluate the model without finetuning on VideoAttentionTarget, ChildPlay, and GOO-Real \cite{tomas2021goo}. As shown in Table~\ref{tab:cross_dataset}, OmniGF achieves the lowest Dist. on all three unseen datasets and the highest PLAH on VideoAttentionTarget and ChildPlay. Consistent with our in-domain evaluation, AUC for this setting is reported in Appendix~\ref{sec:appendix_auc_plah}.

\begin{table}[t]
\centering
\small
\caption{Cross-dataset evaluation. All models are trained on GazeFollow and evaluated on unseen datasets without fine-tuning. PLAH is not computed on GOO-Real as it does not contain human heads as gaze targets.}
\label{tab:cross_dataset}
\begin{tabular}{lccccc}
\toprule
\multirow{2}{*}{Method} & \multicolumn{2}{c}{VideoAttentionTarget} & \multicolumn{2}{c}{ChildPlay} & GOO-Real \\
\cmidrule(lr){2-3} \cmidrule(lr){4-5} \cmidrule(lr){6-6}
& Dist.$\downarrow$ & PLAH$\uparrow$ & Dist.$\downarrow$ & PLAH$\uparrow$ & Dist.$\downarrow$ \\
\midrule
Chong \cite{chong2020detecting}            & 0.119 & -- & 0.121 & -- & 0.334 \\
Jin \cite{jin2022depth}                    & 0.104 & -- & -- & -- & -- \\ 
Tonini w/ UDA \cite{tonini2022multimodal}  & -- & -- & -- & -- & 0.238 \\
Miao \cite{miao2023patch}                  & 0.109 & -- & 0.113 & -- & 0.202 \\
Gupta \cite{gupta2022modular}              & 0.137 & -- & 0.142 & -- & -- \\
Tafasca \cite{tafasca2023childplay}        & 0.123 & -- & 0.115 & -- & -- \\
Gaze-LLE (ViT-B) \cite{gazelle}            & 0.105 & 0.713 & 0.114 & 0.505 & \uline{0.174} \\
Gaze-LLE (ViT-L) \cite{gazelle}            & \uline{0.100} & \uline{0.743} & \uline{0.101} & \uline{0.582} & 0.175 \\
\midrule
OmniGF                          & \textbf{0.098} & \textbf{0.772} & \textbf{0.094} & \textbf{0.633} & \textbf{0.166} \\
\bottomrule
\end{tabular}
\end{table}

\subsection{Performance on Semantic Gaze Prediction}

Beyond standard spatial coordinate estimation, our framework seamlessly extends to semantic gaze target recognition by generating an additional ``category'' field within the structured language output. Because our model generates open-vocabulary text rather than discrete class logits, predicted labels may not perfectly align with the predefined test vocabulary. To resolve this, we encode both our text predictions and the evaluation classes using CLIP (utilizing the prompt template ``a photo of a [label]'') and retrieve the ten most similar test classes by cosine similarity. If any of these candidates is a WordNet hypernym of the predicted label, we assign the one closest to it in the hypernym hierarchy; otherwise, we assign the class with the highest similarity.

Furthermore, we identified noticeable noise in the original pseudo-labeled GazeFollow training set established by Semgaze (e.g., a ``cellphone'' being held was incorrectly labeled as ``person''). To ensure high-quality supervision, we refined these pseudo-labels and subsequently trained both our model and a corrected Semgaze$^\ast$ baseline on this improved data. We will publicly release these refined annotations to support future research. Conversely, for GazeHOI, we maintained the original semantic labels because these are derived directly from the ground-truth object bounding boxes of established Human-Object Interaction datasets.

As shown in Tables~\ref{tab:semgaze_gf} and \ref{tab:semgaze_hoi}, OmniGF establishes new state-of-the-art performance on both datasets. On GazeFollow, it achieves a remarkable Acc@1 of 0.649, with a 23.1\% improvement over the corrected SemGaze baseline. Besides, it also excels in both localization and recognition on GazeHOI, outperforming SemGaze by 23.8\%. This confirms that our dual-branch VLM architecture effectively bridges the gap between precise spatial localization and high-level semantic reasoning.


\begin{table}[t]
\centering

\begin{minipage}[t]{0.6\textwidth}
\centering
\caption{Semantic gaze prediction results on GazeFollow. Semgaze$^\ast$ denotes the baseline trained on our refined ground-truth labels for a fair comparison.}
\label{tab:semgaze_gf}
\resizebox{\linewidth}{!}{%
\begin{tabular}{lcccc}
\toprule
\multirow{2}{*}{Method} & \multicolumn{2}{c}{Localization} & \multicolumn{2}{c}{Recognition} \\
\cmidrule(lr){2-3} \cmidrule(lr){4-5}
& Avg. Dist. $\downarrow$ & Min. Dist. $\downarrow$ & Acc@1 $\uparrow$ & MultiAcc@1 $\uparrow$ \\
\midrule
Semgaze \cite{tafasca2024toward}           & 0.108 & 0.051 & 0.447 & 0.516 \\
Semgaze$^\ast$ \cite{tafasca2024toward}     & 0.109 & 0.053 & 0.527 & 0.598 \\
OmniGF                    & \textbf{0.092} & \textbf{0.040} & \textbf{0.649} & \textbf{0.739} \\
\bottomrule
\end{tabular}%
}
\end{minipage}\hfill 
\begin{minipage}[t]{0.38\textwidth}
\centering
\caption{Semantic gaze prediction on GazeHOI. Semgaze* is omitted as semantic labels for GazeHOI are derived directly from existing HOI datasets.}
\label{tab:semgaze_hoi}
\resizebox{\linewidth}{!}{%
\begin{tabular}{lcc}
\toprule
\multirow{2}{*}{Method} & Localization & Recognition \\
\cmidrule(lr){2-2} \cmidrule(lr){3-3} 
 & GazeAcc $\uparrow$ & Acc@1 $\uparrow$ \\
\midrule
Semgaze \cite{tafasca2024toward} & 0.723 & 0.583 \\
OmniGF        & \textbf{0.786} & \textbf{0.722} \\
\bottomrule
\end{tabular}%
}
\end{minipage}

\end{table}

\subsection{Performance on Social Gaze Prediction}

Finally, we extend our framework to the complex task of social gaze prediction using the VSGaze dataset. A unique challenge of this dataset is that it includes additional head bounding boxes and tracks that lack ground-truth gaze target annotations. To effectively train our structured language branch in this dense multi-person environment, we introduce a selective loss masking strategy. Specifically, we input dummy gaze slots for these unannotated individuals and assign them a categorical gaze status of ``unknown.'' During backpropagation, we dynamically exclude the tokens corresponding to these individuals from the language modeling gradient. This ensures the model learns to process all individuals in the scene simultaneously for global context, while focusing its parameter updates strictly on valid, annotated targets.

As shown in Table~\ref{tab:social_gaze}, our framework establishes a new state-of-the-art across all gaze and social reasoning metrics. Crucially, our model performs end-to-end (E2E) social gaze reasoning jointly with gaze localization, eliminating the need for the manual post-processing required by ordinary gaze following methods. Compared to the only E2E baseline MTGS \cite{gupta2024mtgs}, our model improves from 0.590 to 0.742 (25.8\%) on $\text{F1}_{\text{LAEO}}$ and from 0.576 to 0.787 (36.6\%) on $\text{AP}_{\text{SA}}$. These unprecedented gains demonstrate that leveraging the semantic reasoning capabilities of a VLM provides a fundamental advantage in understanding complex social interactions over traditional spatial-only architectures.

\begin{table}[h]
\centering
\small
\caption{Comparison of social gaze reasoning and gaze target estimation performance on VSGaze. E2E denotes end-to-end prediction without relying on manual post-processing heuristics.}
\label{tab:social_gaze}
\begin{tabular}{lcccccc}
\toprule
Method & E2E & Dist. $\downarrow$ & AP$_{\text{IO}}$ $\uparrow$ & F1$_{\text{LAH}}$ $\uparrow$ & F1$_{\text{LAEO}}$ $\uparrow$ & AP$_{\text{SA}}$ $\uparrow$ \\
\midrule
Chong$_S$~\cite{chong2020detecting} & $\times$ & 0.121 & 0.918          & 0.778          & 0.562          & 0.288          \\
Chong$_T$~\cite{chong2020detecting} & $\times$ & 0.130 & 0.956          & 0.764          & 0.529          & 0.331          \\
Gupta~\cite{gupta2022modular}  & $\times$ & 0.119 & 0.929          & 0.784          & 0.590          & 0.335          \\
MTGS~\cite{gupta2024mtgs}               & \checkmark   & 0.107 & 0.940          & 0.795          & 0.590          & 0.576          \\
OmniGF               & \checkmark   & \textbf{0.098} & \textbf{0.992} & \textbf{0.814} & \textbf{0.742} & \textbf{0.787} \\
\bottomrule
\end{tabular}
\end{table}
\section{Conclusion}
\label{sec:conclusion}

In this work, we presented OmniGF, a novel Vision-Language framework for unified gaze following. By introducing a dual-branch architecture that harmonizes a structured language branch with a continuous dense spatial decoder, alongside a dynamic gaze token injection mechanism, we successfully adapted a foundational VLM for high-precision spatial reasoning tasks. OmniGF establishes a new state-of-the-art across diverse benchmarks in a single multi-person forward pass. To the best of our knowledge, this is the first successful application of VLMs to gaze localization that outperforms state-of-the-art vision-based models. Furthermore, benefiting from VLM's strong semantic reasoning capacity, we demonstrate that the model can be seamlessly extended to semantic and social gaze prediction tasks, outperforming current state-of-the-art models by a significant margin. This demonstrates that foundational VLMs can fundamentally unify spatial localization and discrete high-level semantic reasoning. We hope that OmniGF provides a robust, streamlined foundation to accelerate future research in human behavior understanding, social scene analysis, and interactive AI systems.

\clearpage
\section*{Acknowledgements}
This project was partially supported by NSF awards IIS-2123920 and IIS-2212046. Minh Hoai was initially supported by NSF award DUE-2055406, and later in part by the Australian Institute for Machine Learning (Adelaide University) and the Centre for Augmented Reasoning, an initiative of the Australian Government's Department of Education. The authors thank Xiang Li and Jingwei Zhang for helpful discussions.

{
\small
\bibliographystyle{ieeetr}
\bibliography{main}

@string{aaai = "Proceedings of AAAI Conference on Artificial Intelligence"}

@string{accv =  "Proceedings of the Asian Conference on Computer Vision"}

@string{cvpr   = "Proceedings of the IEEE/CVF Conference on Computer Vision and Pattern Recognition"}

@string{cvprw  = "Proceedings of the IEEE/CVF Conference on Computer Vision and Pattern Recognition Workshops"}

@string{eccv  = "Proceedings of the European Conference on Computer Vision"}

@string{fg = "Proceedings of the International Conference on Automatic Face and Gesture Recognition"}

@string{iccv =  "Proceedings of the IEEE/CVF International Conference on Computer Vision"}

@string{icml =  "Proceedings of the International Conference on Machine Learning"}

@string{iros =  "Proceedings of the IEEE/RSJ Conference on Intelligent Robots and Systems"}

@string{nips = "Advances in Neural Information Processing Systems"}

@string{neurips = "Advances in Neural Information Processing Systems"}

@string{pami =  "IEEE Transactions on Pattern Analysis and Machine Intelligence"}

@string{wacv =  "Proceedings of the IEEE/CVF Winter Conference on Applications of Computer Vision"}

@string{it =  "IEEE Information Theory"}

@string{jun = "June"}

@inproceedings{NIPS2015_ec895663,
 author = {Recasens, Adria and Khosla, Aditya and Vondrick, Carl and Torralba, Antonio},
 booktitle = nips,
 editor = {C. Cortes and N. Lawrence and D. Lee and M. Sugiyama and R. Garnett},
 pages = {},
 publisher = {Curran Associates, Inc.},
 title = {Where are they looking?},
 url = {https://proceedings.neurips.cc/paper/2015/file/ec8956637a99787bd197eacd77acce5e-Paper.pdf},
 volume = {28},
 year = {2015}
}

@inproceedings{chong2018connecting,
  title={Connecting gaze, scene, and attention: Generalized attention estimation via joint modeling of gaze and scene saliency},
  author={Chong, Eunji and Ruiz, Nataniel and Wang, Yongxin and Zhang, Yun and Rozga, Agata and Rehg, James M},
  booktitle = eccv,
  pages={383--398},
  year={2018}
}

@inproceedings{fang2021dual,
  title={Dual Attention Guided Gaze Target Detection in the Wild},
  author={Fang, Yi and Tang, Jiapeng and Shen, Wang and Shen, Wei and Gu, Xiao and Song, Li and Zhai, Guangtao},
  booktitle = cvpr,
  pages={11390--11399},
  year={2021}
}

@inproceedings{chong2020detecting,
  title={Detecting attended visual targets in video},
  author={Chong, Eunji and Wang, Yongxin and Ruiz, Nataniel and Rehg, James M},
  booktitle = cvpr,
  pages={5396--5406},
  year={2020}
}

@inproceedings{lian2018believe,
  title={Believe It or Not, We Know What You Are Looking At!},
  author={Lian, Dongze and Yu, Zehao and Gao, Shenghua},
  booktitle = accv,
  pages={35--50},
  year={2018},
  organization={Springer}
}

@InProceedings{Bao_2022_CVPR,
    author    = {Bao, Jun and Liu, Buyu and Yu, Jun},
    title     = {{ESCNet}: Gaze Target Detection With the Understanding of {3D} Scenes},
    booktitle = cvpr,
    month     = {June},
    year      = {2022},
    pages     = {14126--14135}
}

@inproceedings{miao2023patch,
  title={Patch-level Gaze Distribution Prediction for Gaze Following},
  author={Miao, Qiaomu and Hoai, Minh and Samaras, Dimitris},
  booktitle = wacv,
  year={2023}
}

@inproceedings{gupta2022modular,
  title={A Modular Multimodal Architecture for Gaze Target Prediction: Application to Privacy-Sensitive Settings},
  author={Gupta, Anshul and Tafasca, Samy and Odobez, Jean-Marc},
  booktitle = cvprw,
  pages={5041--5050},
  year={2022}
}

@article{hu2022we,
  title={We know where they are looking at from the {RGB-D} camera: Gaze following in {3D}},
  author={Hu, Zhengxi and Yang, Dingye and Cheng, Shilei and Zhou, Lei and Wu, Shichao and Liu, Jingtai},
  journal={IEEE Transactions on Instrumentation and Measurement},
  volume={71},
  pages={1--14},
  year={2022},
  publisher={IEEE}
}

@inproceedings{tafasca2023childplay,
  title={{ChildPlay}: A New Benchmark for Understanding Children's Gaze Behaviour},
  author={Tafasca, Samy and Gupta, Anshul and Odobez, Jean-Marc},
  booktitle = iccv,
  pages={20935--20946},
  year={2023}
}

@inproceedings{tu2022end,
  title={End-to-end human-gaze-target detection with transformers},
  author={Tu, Danyang and Min, Xiongkuo and Duan, Huiyu and Guo, Guodong and Zhai, Guangtao and Shen, Wei},
  booktitle = cvpr,
  pages={2192--2200},
  year={2022},
  organization={IEEE}
}

@inproceedings{tonini2023object,
  title={Object-aware gaze target detection},
  author={Tonini, Francesco and Dall'Asen, Nicola and Beyan, Cigdem and Ricci, Elisa},
  booktitle = iccv,
  pages={21860--21869},
  year={2023}
}

@article{tu2023joint,
  title={Joint gaze-location and gaze-object detection},
  author={Tu, Danyang and Shen, Wei and Sun, Wei and Min, Xiongkuo and Zhai, Guangtao},
  journal={arXiv preprint arXiv:2308.13857},
  year={2023}
}

@inproceedings{tafasca2024sharingan,
  title={Sharingan: A Transformer Architecture for Multi-Person Gaze Following},
  author={Tafasca, Samy and Gupta, Anshul and Odobez, Jean-Marc},
  booktitle = cvpr,
  pages={2008--2017},
  year={2024}
}

@inproceedings{multimae,
  title={{MultiMAE}: Multi-modal multi-task masked autoencoders},
  author={Bachmann, Roman and Mizrahi, David and Atanov, Andrei and Zamir, Amir},
  booktitle = eccv,
  pages={348--367},
  year={2022},
  organization={Springer}
}

@inproceedings{gazelle,
  title={{Gaze-LLE}: Gaze target estimation via large-scale learned encoders},
  author={Ryan, Fiona and Bati, Ajay and Lee, Sangmin and Bolya, Daniel and Hoffman, Judy and Rehg, James M},
  booktitle={Proceedings of the Computer Vision and Pattern Recognition Conference},
  pages={28874--28884},
  year={2025}
}

@article{oquab2023dinov2,
  title={{DINOv2}: Learning robust visual features without supervision},
  author={Oquab, Maxime and Darcet, Timoth{\'e}e and Moutakanni, Th{\'e}o and Vo, Huy and Szafraniec, Marc and Khalidov, Vasil and Fernandez, Pierre and Haziza, Daniel and Massa, Francisco and El-Nouby, Alaaeldin and others},
  journal={Transactions on Machine Learning Research},
  year={2024}
}

@inproceedings{tafasca2024toward,
  title={Toward semantic gaze target detection},
  author={Tafasca, Samy and Gupta, Anshul and Bros, Victor and Odobez, Jean-Marc},
  booktitle=neurips,
  volume={37},
  pages={121422--121448},
  year={2024}
}

@inproceedings{gupta2024mtgs,
title={{MTGS}: A Novel Framework for Multi-Person Temporal Gaze Following and Social Gaze Prediction},
author={Anshul Gupta and Samy Tafasca and Arya Farkhondeh and Pierre Vuillecard and Jean-Marc Odobez},
booktitle = neurips,
year={2024},
url={https://openreview.net/forum?id=ALU676zGFE}
}

@inproceedings{miao2025multi,
  title={Multi-view Gaze Target Estimation},
  author={Miao, Qiaomu and Golani, Vivek Raju and Xu, Jingyi and Dutta, Progga Paromita and Hoai, Minh and Samaras, Dimitris},
  booktitle={Proceedings of the IEEE/CVF International Conference on Computer Vision},
  year={2025}
}

@inproceedings{yang2024aaai,
  title={Gaze Target Detection by Merging Human Attention and Activity Cues},
  author={Yang, Yaokun and Yin, Yihan and Lu, Feng},
  booktitle = aaai,
  pages={6585--6593},
  year={2024}
}

@inproceedings{hou2024multi,
  title={Multi-modal gaze following in conversational scenarios},
  author={Hou, Yuqi and Zhang, Zhongqun and Horanyi, Nora and Moon, Jaewon and Cheng, Yihua and Chang, Hyung Jin},
  booktitle={Proceedings of the IEEE/CVF Winter Conference on Applications of Computer Vision},
  pages={1186--1195},
  year={2024}
}

@InProceedings{liu2023visual,
  title = {Visual Instruction Tuning},
  author = {Liu, Haotian and Li, Chunyuan and Wu, Qingyang and Lee, Yong Jae},
  booktitle = {Advances in Neural Information Processing Systems},
  volume = {36},
  year = {2023}
}

@article{bai2023qwen,
  title = {{Qwen-VL}: A Versatile Vision-Language Model for Understanding, Localization, Text Reading, and Beyond},
  author = {Bai, Jinze and Bai, Shuai and Yang, Shusheng and Wang, Shijie and Tan, Sinan and Wang, Peng and Lin, Junyang and Zhou, Chang and Zhou, Jingren},
  journal = {arXiv preprint arXiv:2308.12966},
  year = {2023}
}

@article{chen2023shikra,
  title = {Shikra: Unleashing Multimodal {LLM}'s Referential Dialogue Magic},
  author = {Chen, Keqin and Zhang, Zhao and Zeng, Weili and Zhang, Richong and Zhu, Feng and Zhao, Rui},
  journal = {arXiv preprint arXiv:2306.15195},
  year = {2023}
}

@InProceedings{rasheed2024glamm,
  title = {{GLaMM}: Pixel Grounding Large Multimodal Model},
  author = {Rasheed, Hanoona and Maaz, Muhammad and Shaji, Sahal and Shaker, Abdelrahman and Khan, Salman and Cholakkal, Hisham and Anwer, Rao M. and Xing, Eric and Yang, Ming-Hsuan and Khan, Fahad S.},
  booktitle = {Proceedings of the IEEE/CVF Conference on Computer Vision and Pattern Recognition (CVPR)},
  year = {2024},
  month = {June}
}

@article{beyer2024paligemma,
  title = {{PaliGemma}: A Versatile {3B} {VLM} for Transfer},
  author = {Beyer, Lucas and Steiner, Andreas and Pinto, Andr{\'e} Susano and Kolesnikov, Alexander and Wang, Xiao and Salz, Daniel and Neumann, Maxim and Alabdulmohsin, Ibrahim and Tschannen, Michael and Bugliarello, Emanuele and Unterthiner, Thomas and Keysers, Daniel and Koppula, Skanda and Liu, Fangyu and Grycner, Adam and Gritsenko, Alexey and Houlsby, Neil and Kumar, Manoj and Rong, Keran and Eisenschlos, Julian and Kabra, Rishabh and Bauer, Matthias and Bo{\v{s}}njak, Matko and Chen, Xi and Minderer, Matthias and Voigtlaender, Paul and Bica, Ioana and Balazevic, Ivana and Puigcerver, Joan and Papalampidi, Pinelopi and Henaff, Olivier and Xiong, Xi and Soricut, Radu and Harmsen, Jeremiah and Zhai, Xiaohua},
  journal = {arXiv preprint arXiv:2407.07726},
  year = {2024}
}

@InProceedings{glip,
  title = {Grounded Language-Image Pre-training},
  author = {Li, Liunian Harold and Zhang, Pengchuan and Zhang, Haotian and Yang, Jianwei and Li, Chunyuan and Zhong, Yiwu and Wang, Lijuan and Yuan, Lu and Zhang, Lei and Hwang, Jenq-Neng and Chang, Kai-Wei and Gao, Jianfeng},
  booktitle = {Proceedings of the IEEE/CVF Conference on Computer Vision and Pattern Recognition (CVPR)},
  year = {2022}
}

@InProceedings{lai2024lisa,
  title = {{LISA}: Reasoning Segmentation via Large Language Model},
  author = {Lai, Xin and Tian, Zhuotao and Chen, Yukang and Li, Yanwei and Yuan, Yuhui and Liu, Shu and Jia, Jiaya},
  booktitle = {Proceedings of the IEEE/CVF Conference on Computer Vision and Pattern Recognition (CVPR)},
  pages = {9579--9589},
  year = {2024},
  month = {June}
}

@InProceedings{gupta2024exploring,
  title = {Exploring the Zero-Shot Capabilities of Vision-Language Models for Improving Gaze Following},
  author = {Gupta, Anshul and Vuillecard, Pierre and Farkhondeh, Arya and Odobez, Jean-Marc},
  booktitle = {Proceedings of the IEEE/CVF Conference on Computer Vision and Pattern Recognition Workshops (CVPRW)},
  year = {2024}
}

@InProceedings{miao2024diffusion,
  title = {Diffusion-Refined {VQA} Annotations for Semi-Supervised Gaze Following},
  author = {Miao, Qiaomu and Graikos, Alexandros and Zhang, Jingwei and Mondal, Sounak and Hoai, Minh and Samaras, Dimitris},
  booktitle = {European Conference on Computer Vision},
  year = {2024}
}

@article{mathew2025gazevlm,
  title = {{GazeVLM}: A Vision-Language Model for Multi-Task Gaze Understanding},
  author = {Mathew, Athul M. and Hermassi, Haithem and Khalid, Thariq and Khan, Arshad Ali},
  journal = {arXiv preprint arXiv:2511.06348},
  year = {2025}
}

@article{wang2025vl4gaze,
  title = {{VL4Gaze}: Unleashing Vision-Language Models for Gaze Following},
  author = {Wang, Shijing and Cui, Chaoqun and Huang, Yaping and Chang, Hyung Jin and Cheng, Yihua},
  journal = {arXiv preprint arXiv:2512.20735},
  year = {2025}
}

@article{wang2024qwen2,
  title={{Qwen2-VL}: Enhancing vision-language model's perception of the world at any resolution},
  author={Wang, Peng and Bai, Shuai and Tan, Sinan and Wang, Shijie and Fan, Zhihao and Bai, Jinze and Chen, Keqin and Liu, Xuejing and Wang, Jialin and Ge, Wenbin and others},
  journal={arXiv preprint arXiv:2409.12191},
  year={2024}
}

@article{bai2025qwen3,
  title={{Qwen3-VL} technical report},
  author={Bai, Shuai and Cai, Yuxuan and Chen, Ruizhe and Chen, Keqin and Chen, Xionghui and Cheng, Zesen and Deng, Lianghao and Ding, Wei and Gao, Chang and Ge, Chunjiang and others},
  journal={arXiv preprint arXiv:2511.21631},
  year={2025}
}

@inproceedings{resnet,
  title={Deep residual learning for image recognition},
  author={He, Kaiming and Zhang, Xiangyu and Ren, Shaoqing and Sun, Jian},
  booktitle = cvpr,
  pages={770--778},
  year={2016}
}

@inproceedings{gaze360,
  title={Gaze360: Physically unconstrained gaze estimation in the wild},
  author={Kellnhofer, Petr and Recasens, Adria and Stent, Simon and Matusik, Wojciech and Torralba, Antonio},
  booktitle = iccv,
  pages={6912--6921},
  year={2019}
}

@inproceedings{hu2022lora,
  title={{LoRA}: Low-Rank Adaptation of Large Language Models},
  author={Hu, Edward J and Shen, Yelong and Wallis, Phillip and Allen-Zhu, Zeyuan and Li, Yuanzhi and Wang, Shean and Wang, Lu and Chen, Weizhu},
  booktitle={International Conference on Learning Representations},
  year={2022}
}

@inproceedings{jin2021multi,
  title={Multi-person gaze-following with numerical coordinate regression},
  author={Jin, Tianlei and Lin, Zheyuan and Zhu, Shiqiang and Wang, Wen and Hu, Shunda},
  booktitle={2021 16th IEEE International Conference on Automatic Face and Gesture Recognition (FG 2021)},
  pages={1--8},
  year={2021},
  organization={IEEE}
}

@article{jin2022depth,
  title={Depth-aware gaze-following via auxiliary networks for robotics},
  author={Jin, Tianlei and Yu, Qizhi and Zhu, Shiqiang and Lin, Zheyuan and Ren, Jie and Zhou, Yuanhai and Song, Wei},
  journal={Engineering Applications of Artificial Intelligence},
  volume={113},
  pages={104924},
  year={2022},
  publisher={Elsevier}
}

@inproceedings{fan2018inferring,
  title={Inferring shared attention in social scene videos},
  author={Fan, Lifeng and Chen, Yixin and Wei, Ping and Wang, Wenguan and Zhu, Song-Chun},
  booktitle=cvpr,
  pages={6460--6468},
  year={2018}
}

@inproceedings{marin2019laeo,
  title={{LAEO-Net}: revisiting people looking at each other in videos},
  author={Marin-Jimenez, Manuel J and Kalogeiton, Vicky and Medina-Suarez, Pablo and Zisserman, Andrew},
  booktitle=cvpr,
  pages={3477--3485},
  year={2019}
}

@inproceedings{tonini2022multimodal,
  title={Multimodal across domains gaze target detection},
  author={Tonini, Francesco and Beyan, Cigdem and Ricci, Elisa},
  booktitle={Proceedings of the 2022 International Conference on Multimodal Interaction},
  pages={420--431},
  year={2022}
}

@article{sheikhi2015combining,
  title={Combining dynamic head pose--gaze mapping with the robot conversational state for attention recognition in human--robot interactions},
  author={Sheikhi, Samira and Odobez, Jean-Marc},
  journal={Pattern Recognition Letters},
  volume={66},
  pages={81--90},
  year={2015},
  publisher={Elsevier}
}

@article{emery2000eyes,
  title={The eyes have it: the neuroethology, function and evolution of social gaze},
  author={Emery, Nathan J},
  journal={Neuroscience \& Biobehavioral Reviews},
  volume={24},
  number={6},
  pages={581--604},
  year={2000},
  publisher={Elsevier}
}

@article{admoni2017social,
  title={Social eye gaze in human-robot interaction: a review},
  author={Admoni, Henny and Scassellati, Brian},
  journal={Journal of Human-Robot Interaction},
  volume={6},
  number={1},
  pages={25--63},
  year={2017},
  publisher={Journal of Human-Robot Interaction Steering Committee}
}

@article{senju2009atypical,
  title={Atypical eye contact in autism: Models, mechanisms and development},
  author={Senju, Atsushi and Johnson, Mark H},
  journal={Neuroscience \& Biobehavioral Reviews},
  volume={33},
  number={8},
  pages={1204--1214},
  year={2009},
  publisher={Elsevier}
}

@inproceedings{kasahara2022look,
  title={Look both ways: Self-supervising driver gaze estimation and road scene saliency},
  author={Kasahara, Isaac and Stent, Simon and Park, Hyun Soo},
  booktitle={European Conference on Computer Vision},
  pages={126--142},
  year={2022},
  organization={Springer}
}

@article{de2020computer,
  title={Computer vision in autism spectrum disorder research: a systematic review of published studies from 2009 to 2019},
  author={De Belen, Ryan Anthony J and Bednarz, Tomasz and Sowmya, Arcot and Del Favero, Dennis},
  journal={Translational Psychiatry},
  volume={10},
  number={1},
  pages={333},
  year={2020},
  publisher={Nature Publishing Group UK London}
}

@article{masse2017tracking,
  title={Tracking gaze and visual focus of attention of people involved in social interaction},
  author={Mass{\'e}, Beno{\^\i}t and Ba, Sil{\`e}ye and Horaud, Radu},
  journal=pami,
  volume={40},
  number={11},
  pages={2711--2724},
  year={2018},
  publisher={IEEE}
}

@inproceedings{saran2018human,
  title={Human gaze following for human-robot interaction},
  author={Saran, Akanksha and Majumdar, Srinjoy and Short, Elaine Schaertl and Thomaz, Andrea and Niekum, Scott},
  booktitle={2018 IEEE/RSJ International Conference on Intelligent Robots and Systems (IROS)},
  pages={8615--8621},
  year={2018},
  organization={IEEE}
}

@inproceedings{liu2019gaze,
  title={A gaze model improves autonomous driving},
  author={Liu, Congcong and Chen, Yuying and Tai, Lei and Ye, Haoyang and Liu, Ming and Shi, Bertram E},
  booktitle={Proceedings of the 11th ACM Symposium on Eye Tracking Research \& Applications},
  pages={1--5},
  year={2019}
}

@article{martin2018dynamics,
  title={Dynamics of driver's gaze: Explorations in behavior modeling and maneuver prediction},
  author={Martin, Sujitha and Vora, Sourabh and Yuen, Kevan and Trivedi, Mohan Manubhai},
  journal={IEEE Transactions on Intelligent Vehicles},
  volume={3},
  number={2},
  pages={141--150},
  year={2018},
  publisher={IEEE}
}

@inproceedings{cao2026gaze,
  title={Gaze Target Estimation Anywhere with Concepts},
  author={Cao, Xu and Yang, Houze and Gunda, Vipin and Zhou, Zhongyi and Xu, Tianyu and Kowdle, Adarsh and Kim, Inki and Rehg, James M.},
  booktitle = cvpr,
  year={2026}
}

@inproceedings{tomas2021goo,
  title={{GOO}: A Dataset for Gaze Object Prediction in Retail Environments},
  author={Tomas, Henri and Reyes, Marcus and Dionido, Raimarc and Ty, Mark and Mirando, Jonric and Casimiro, Joel and Atienza, Rowel and Guinto, Richard},
  booktitle = cvprw,
  pages={3119--3127},
  year={2021}
}

@article{wang2025internvl3_5,
  title={{InternVL3.5}: Advancing Open-Source Multimodal Models in Versatility, Reasoning, and Efficiency},
  author={Wang, Weiyun and Gao, Zhangwei and others},
  journal={arXiv preprint arXiv:2508.18265},
  year={2025}
}

@misc{gemini36flash,
  title={{Gemini 3.6 Flash}},
  author={{Google DeepMind}},
  howpublished={\url{https://aistudio.google.com/docs/models/gemini-3.6-flash}},
  year={2026}
}

@inproceedings{radford2021learning,
  title={Learning Transferable Visual Models From Natural Language Supervision},
  author={Radford, Alec and Kim, Jong Wook and Hallacy, Chris and Ramesh, Aditya and Goh, Gabriel and Agarwal, Sandhini and Sastry, Girish and Askell, Amanda and Mishkin, Pamela and Clark, Jack and Krueger, Gretchen and Sutskever, Ilya},
  booktitle=icml,
  pages={8748--8763},
  year={2021}
}

@inproceedings{yin2024clipgaze,
  title={{CLIP-Gaze}: Towards General Gaze Estimation via Visual-Linguistic Model},
  author={Yin, Pengwei and Zeng, Guanzhong and Wang, Jingjing and Xie, Di},
  booktitle=aaai,
  pages={6729--6737},
  year={2024}
}

@article{wang2026vlhtr,
  title={{VL-HTR}: Learning Human--Target Representation From Vision--Language Model},
  author={Wang, Binglu and Guo, Chenxi and Cui, Jingyi and Xia, Haisheng and Guo, Guangyu and Li, Zhijun},
  journal={IEEE Transactions on Cybernetics},
  volume={56},
  number={5},
  pages={2638--2649},
  year={2026}
}

@article{wang2026eyesonvlm,
  title={Eyes on {VLM}: Benchmarking Gaze Following and Social Gaze Prediction in Vision Language Models},
  author={Wang, Hengfei and Gupta, Anshul and Vuillecard, Pierre and Odobez, Jean-Marc},
  journal={arXiv preprint arXiv:2605.19859},
  year={2026}
}
}

\newpage
\appendix

\section{System Prompts}
\label{sec:appendix_prompts}

To effectively guide the Vision-Language Model (VLM) backbone to generate structured reasoning states, we utilize specific system prompts during both training and inference. These prompts inform the model of the input format, establish a normalized coordinate space, and strictly dictate the JSON schema of the desired output.

\textbf{Basic Gaze Target Estimation.} For standard spatial gaze target estimation, we use the following base system prompt:

\begin{quote}
\small
\textit{"You are given an image with head bounding boxes for all people in the scene. Each person is identified as P0..PN with their head box given as \texttt{\{"bbox\_2d": [x1, y1, x2, y2]\}} in coordinates [0, 999]. For each person, predict whether their gaze target is inside or outside the image. Output one JSON object per person. If the gaze target is inside, output \texttt{\{"status": "inside", "point\_2d": [x, y]\}}. If outside, output \texttt{\{"status": "outside"\}}."}
\end{quote}

\textbf{Semantic Gaze Prediction.} For the semantic gaze target recognition task (evaluated on GazeFollow and GazeHOI), we extend the base prompt to simultaneously reason for the gaze target semantics. We achieve this by appending an instruction to predict what the person is looking at and updating the required JSON schema for in-frame targets. The full system prompt is formulated as follows:

\begin{quote}
\small
\textit{"You are given an image with head bounding boxes for all people in the scene. Each person is identified as P0..PN with their head box given as \texttt{\{"bbox\_2d": [x1, y1, x2, y2]\}} in coordinates [0, 999]. For each person, predict whether their gaze target is inside or outside the image, and what they are looking at. Output one JSON object per person. If the gaze target is inside, output \texttt{\{"status": "inside", "point\_2d": [x, y], "category": "<object>"\}}. If outside, output \texttt{\{"status": "outside"\}}."}
\end{quote}

\section{Additional Implementation Details}
\label{sec:appendix_impl}

All models use the Qwen3-VL-4B-Instruct backbone with LoRA ($r=16$, $\alpha=32$) applied to the query and value projections of all 36 transformer layers of the language model, a Gaze360-pretrained ResNet-18 head encoder \cite{gaze360}, $448 \times 448$ scene inputs, $224 \times 224$ head crops, and $64 \times 64$ gaze heatmaps per person. All models are trained on a single GPU with AdamW, a batch size of 8, and two gradient accumulation steps (effective batch size 16), with loss coefficients $\lambda_{lm} = 1.0$ and $\lambda_{hm} = 10.0$, and $\lambda_{soc} = 1.0$ for social gaze prediction. One learning rate is used for the LoRA parameters and another for the head encoder and all prediction heads.

On GazeFollow, one epoch covers 103,455 training images, i.e., 6,466 optimizer steps, and 20 epochs amount to 129,320 steps. We train two GazeFollow models: the model reported in Table~\ref{tab:gf_vat_results}, trained on the standard annotations with $\lambda_{inout} = 0$, and a model trained with in/out supervision ($\lambda_{inout} = 0.1$) on training annotations that additionally include out-of-frame persons, which serves as the initialization for VideoAttentionTarget, ChildPlay, and VSGaze. For every task, we select the checkpoint with the best validation metric (Dist. for gaze localization, Acc@1 for semantic gaze prediction). On VideoAttentionTarget and ChildPlay, the best validation distance is reached after the first epoch, i.e., after 568 and 1,037 optimizer steps, respectively. Table~\ref{tab:supp_impl} summarizes the per-task settings.

\begin{table}[h]
\centering
\small
\caption{Per-task training settings. LR is given as LoRA / head encoder and prediction heads. Cosine ($T$) denotes one epoch of linear warmup followed by cosine annealing over $T$ epochs.}
\label{tab:supp_impl}
\setlength{\tabcolsep}{4pt}
\begin{tabular}{lcc}
\toprule
Task / dataset & LR & Schedule \\
\midrule
GazeFollow           & $10^{-4}$ / $2.5{\times}10^{-4}$ & Cosine (20) \\
VideoAttentionTarget & $2.5{\times}10^{-5}$ / $5{\times}10^{-5}$ & Constant \\
ChildPlay            & $2.5{\times}10^{-5}$ / $2.5{\times}10^{-5}$ & Cosine (3) \\
Semantic, GazeFollow & $5{\times}10^{-5}$ / $10^{-4}$ & Cosine (15) \\
Semantic, GazeHOI    & $5{\times}10^{-5}$ / $10^{-4}$ & Constant \\
Social, VSGaze       & $5{\times}10^{-5}$ / $5{\times}10^{-5}$ & Constant \\
\bottomrule
\end{tabular}
\end{table}

\section{Additional Evaluation Metrics: AUC and PLAH}
\label{sec:appendix_auc_plah}

In the main paper, we report AUC only on GazeFollow. AUC has known limitations in gaze following evaluation. For single-annotation benchmarks such as VideoAttentionTarget and ChildPlay, the ground-truth heatmap is an artificially constructed Gaussian with a predefined radius centered on the single annotated gaze point, so AUC merely measures similarity to this synthetic target and is largely uninformative \cite{tafasca2024sharingan}. Consequently, several recent methods have phased out AUC in favor of distance-based metrics, either for single-annotation benchmarks \cite{tafasca2024sharingan, cao2026gaze, gupta2024mtgs} or for all gaze following benchmarks \cite{yang2024aaai}. For completeness, we report AUC on all benchmarks, together with the looking-at-heads precision (PLAH) used by \cite{tafasca2024sharingan}, in both the in-domain (Table~\ref{tab:supp_auc_plah}) and cross-dataset (Table~\ref{tab:supp_cross_auc_plah}) settings. We follow the PLAH definition of \cite{tafasca2024sharingan}, under which a prediction is counted as positive only if the predicted and annotated gaze points fall within the same head box. PLAH values for Tafasca et al.~\cite{tafasca2023childplay} and Sharingan are taken from \cite{tafasca2024sharingan}. PLAH for Gaze-LLE was computed by us with the same evaluation code as for OmniGF, using the official checkpoints released by the authors. As GOO-Real does not contain human heads as gaze targets, PLAH is not computed on it.

Although OmniGF does not achieve the highest AUC, it achieves the best distance and PLAH scores on all datasets in both settings. We maintain that distance metrics, which directly evaluate the geometric precision of the predicted gaze target, remain the most reliable indicator of localization performance for this task.

\begin{table}[h]
\centering
\footnotesize
\caption{In-domain comparison including AUC and PLAH on GazeFollow (GF), VideoAttentionTarget (VAT), and ChildPlay (CP). Avg and Min denote the Avg Dist. and Min Dist. on GazeFollow. Best numbers are marked as bold while second-best are underlined.}
\label{tab:supp_auc_plah}
\setlength{\tabcolsep}{2.5pt}
\begin{tabular}{@{}lcccccccccccc@{}}
\toprule
\multirow{2}{*}{Method} & \multicolumn{4}{c}{GazeFollow} & \multicolumn{4}{c}{VideoAttentionTarget} & \multicolumn{4}{c}{ChildPlay} \\
\cmidrule(lr){2-5} \cmidrule(lr){6-9} \cmidrule(lr){10-13}
& Avg$\downarrow$ & Min$\downarrow$ & AUC$\uparrow$ & PLAH$\uparrow$ & Dist.$\downarrow$ & AUC$\uparrow$ & AP$\uparrow$ & PLAH$\uparrow$ & Dist.$\downarrow$ & AUC$\uparrow$ & AP$\uparrow$ & PLAH$\uparrow$ \\
\midrule
Chong \cite{chong2020detecting}          & 0.137 & 0.077 & 0.921 & -- & 0.134 & 0.860 & 0.853 & -- & -- & -- & -- & -- \\
Miao \cite{miao2023patch}                & 0.123 & 0.065 & 0.934 & -- & 0.109 & 0.917 & \uline{0.908} & -- & -- & -- & -- & -- \\
Gupta \cite{gupta2022modular}            & 0.114 & 0.056 & 0.943 & -- & 0.110 & -- & 0.879 & -- & 0.113 & 0.919 & 0.983 & -- \\
Tafasca \cite{tafasca2023childplay}      & 0.125 & 0.064 & 0.936 & 0.622 & 0.109 & -- & 0.834 & \uline{0.752} & 0.107 & 0.935 & 0.986 & 0.590 \\
Sharingan \cite{tafasca2024sharingan}    & 0.113 & 0.057 & 0.944 & 0.667 & 0.107 & -- & 0.891 & 0.738 & 0.106 & -- & 0.990 & 0.600 \\
MTGS \cite{gupta2024mtgs}                & 0.116 & 0.059 & 0.929 & -- & 0.105 & -- & 0.869 & -- & 0.117 & -- & 0.993 & -- \\
Gaze-LLE (ViT-B) \cite{gazelle}          & 0.104 & 0.045 & \uline{0.956} & 0.665 & 0.107 & 0.933 & 0.897 & 0.722 & 0.106 & \uline{0.949} & \uline{0.994} & 0.579 \\
Gaze-LLE (ViT-L) \cite{gazelle}          & \uline{0.099} & \uline{0.041} & \textbf{0.958} & \uline{0.679} & 0.103 & \uline{0.937} & 0.903 & 0.738 & \uline{0.101} & \textbf{0.951} & \uline{0.994} & \uline{0.612} \\
GazeVLM \cite{mathew2025gazevlm}         & 0.123 & 0.061 & 0.936 & -- & \uline{0.102} & \textbf{0.948} & 0.898 & -- & -- & -- & -- & -- \\
\midrule
OmniGF (proposed)                        & \textbf{0.091} & \textbf{0.040} & 0.941 & \textbf{0.753} & \textbf{0.096} & 0.917 & \textbf{0.923} & \textbf{0.759} & \textbf{0.090} & 0.939 & \textbf{0.996} & \textbf{0.662} \\
\bottomrule
\end{tabular}
\end{table}

\begin{table}[h]
\centering
\small
\caption{Cross-dataset evaluation including AUC and PLAH. All models are trained on GazeFollow and evaluated on unseen datasets without fine-tuning. PLAH is not computed on GOO-Real as it does not contain human heads as gaze targets.}
\label{tab:supp_cross_auc_plah}
\setlength{\tabcolsep}{3pt}
\begin{tabular}{lcccccccc}
\toprule
\multirow{2}{*}{Method} & \multicolumn{3}{c}{VideoAttentionTarget} & \multicolumn{3}{c}{ChildPlay} & \multicolumn{2}{c}{GOO-Real} \\
\cmidrule(lr){2-4} \cmidrule(lr){5-7} \cmidrule(lr){8-9}
& Dist.$\downarrow$ & AUC$\uparrow$ & PLAH$\uparrow$ & Dist.$\downarrow$ & AUC$\uparrow$ & PLAH$\uparrow$ & Dist.$\downarrow$ & AUC$\uparrow$ \\
\midrule
Chong \cite{chong2020detecting}            & 0.119 & 0.906 & -- & 0.121 & 0.912 & -- & 0.334 & 0.670 \\
Jin \cite{jin2022depth}                    & 0.104 & 0.900 & -- & -- & -- & -- & -- & -- \\ 
Tonini w/ UDA \cite{tonini2022multimodal}  & -- & -- & -- & -- & -- & -- & 0.238 & 0.840 \\
Miao \cite{miao2023patch}                  & 0.109 & 0.923 & -- & 0.113 & 0.933 & -- & 0.202 & 0.869 \\
Gupta \cite{gupta2022modular}              & 0.137 & 0.907 & -- & 0.142 & 0.923 & -- & -- & -- \\
Tafasca \cite{tafasca2023childplay}        & 0.123 & 0.911 & -- & 0.115 & 0.932 & -- & -- & -- \\
Gaze-LLE (ViT-B) \cite{gazelle}            & 0.105 & \uline{0.932} & 0.713 & 0.114 & \uline{0.946} & 0.505 & \uline{0.174} & \textbf{0.901} \\
Gaze-LLE (ViT-L) \cite{gazelle}            & \uline{0.100} & \textbf{0.937} & \uline{0.743} & \uline{0.101} & \textbf{0.951} & \uline{0.582} & 0.175 & \uline{0.898} \\
\midrule
OmniGF (proposed)                          & \textbf{0.098} & 0.908 & \textbf{0.772} & \textbf{0.094} & 0.933 & \textbf{0.633} & \textbf{0.166} & 0.861 \\
\bottomrule
\end{tabular}
\end{table}

\section{Ablation on Person Token Ordering}
\label{sec:appendix_person_order}

Since OmniGF processes all individuals in a scene jointly through a single structured prompt, the person tokens $(P_1, P_2, \dots, P_N)$ must be presented to the model in some order. During training, we randomly shuffle the person order. In all other experiments, persons are presented at inference time in their default order in the dataset annotation files, without enforcing any specific spatial or logical ordering. To assess the sensitivity of the model to this sequence, we evaluate several person-ordering strategies at inference time on VideoAttentionTarget and ChildPlay, which naturally contain multiple people per image.

As shown in Table~\ref{tab:supp_person_order}, varying the person ordering results in only negligible performance fluctuations (at most 0.004 in Dist. and AP). This confirms that OmniGF does not rely on a specific sequence of person tokens and remains robust to the order in which individuals are presented.

\begin{table}[h]
\centering
\small
\caption{Sensitivity to person-prompt ordering at inference time. LR--TB and RL--BT denote left-to-right/top-to-bottom and right-to-left/bottom-to-top spatial ordering, respectively. Random denotes the average over 3 random permutations with different seeds. Default denotes the order in the dataset annotation files, used in all other experiments.}
\label{tab:supp_person_order}
\begin{tabular}{lcccc}
\toprule
\multirow{2}{*}{Person ordering} & \multicolumn{2}{c}{ChildPlay} & \multicolumn{2}{c}{VideoAttentionTarget} \\
\cmidrule(lr){2-3} \cmidrule(lr){4-5}
& Dist.$\downarrow$ & AP$\uparrow$ & Dist.$\downarrow$ & AP$\uparrow$ \\
\midrule
LR--TB  & 0.090 & 0.997 & 0.097 & 0.923 \\
RL--BT  & 0.093 & 0.996 & 0.100 & 0.919 \\
Random  & 0.090 & 0.996 & 0.098 & 0.919 \\
Default & 0.090 & 0.996 & 0.096 & 0.923 \\
\bottomrule
\end{tabular}
\end{table}


\section{Ablation on Architectural Design Choices}
\label{sec:appendix_arch_ablation}

To complement the component-level ablation in the main paper (Table~\ref{tab:ablation_components}), we evaluate finer-grained design choices along three axes: the position from which the person token is extracted, the strategy for injecting head embeddings, and the backbone layer from which hidden states are taken for the gaze branch. For each experiment, we vary one component at a time while keeping the rest of the model and training configuration unchanged, and train and evaluate on GazeFollow.

As shown in Table~\ref{tab:supp_arch_ablation}, using the first generated token of each person's output as the person token significantly outperforms both mean-pooling over the person's generated tokens and using the last generated token. We hypothesize that the first token acts as an anchor that cleanly captures the initial reasoning state for that individual, before it is mixed with the information needed to generate the remaining text tokens. For head token injection, replacing the dedicated \texttt{<|gaze\_pad|>} embedding with the head embedding yields better localization than adding the head embedding to the embedding of the person identifier token ($P_i$) in the input prompt. Regarding hidden layer selection, a multi-layer DPT-style fusion of four intermediate layers offers a marginal improvement in AUC, whereas taking the hidden states from the final layer provides the best Avg Dist. and Min Dist. without the computational overhead of fusing multiple high-resolution feature maps.

\begin{table}[h]
\centering
\small
\caption{Ablation on architectural design choices on GazeFollow. Each block varies one component while keeping the others fixed. Our final design is shaded.}
\label{tab:supp_arch_ablation}
\begin{tabular}{llccc}
\toprule
Axis & Variant & Avg Dist.$\downarrow$ & Min Dist.$\downarrow$ & AUC$\uparrow$ \\
\midrule
\multirow{3}{*}{Person token}
 & Last generated token & 0.288 & 0.216 & 0.566 \\
 & Mean-pooled generated tokens & 0.104 & 0.047 & 0.852 \\
 & \cellcolor{gray!15}First generated token & \cellcolor{gray!15}0.091 & \cellcolor{gray!15}0.040 & \cellcolor{gray!15}0.941 \\
\midrule
\multirow{2}{*}{Head injection}
 & Add to $P_i$ token embedding & 0.096 & 0.042 & 0.939 \\
 & \cellcolor{gray!15}Replace \texttt{<|gaze\_pad|>} embedding & \cellcolor{gray!15}0.091 & \cellcolor{gray!15}0.040 & \cellcolor{gray!15}0.941 \\
\midrule
\multirow{4}{*}{Hidden layer}
 & Layer 9 (early) & 0.103 & 0.044 & 0.940 \\
 & Layer 18 (middle) & 0.095 & 0.041 & 0.943 \\
 & DPT-style fusion of 4 layers & 0.094 & 0.042 & 0.944 \\
 & \cellcolor{gray!15}Final layer (36) & \cellcolor{gray!15}0.091 & \cellcolor{gray!15}0.040 & \cellcolor{gray!15}0.941 \\
\bottomrule
\end{tabular}
\end{table}

\section{Zero-shot Performance of Other VLMs}
\label{sec:appendix_zeroshot_vlm}

In the ablation study (Table~\ref{tab:ablation_components}), we report the zero-shot performance of the Qwen3-VL-4B-Instruct backbone. To examine whether this limitation persists with larger or different foundation models, we further evaluate Qwen3-VL at multiple scales (2B, 4B, 8B, and 32B), InternVL3.5-8B, and the commercial model Gemini-3.6-Flash on the GazeFollow test set in a zero-shot setting. All models are prompted with the same system and person prompts used by OmniGF (Appendix~\ref{sec:appendix_prompts}) to generate the gaze status and target coordinates for every person in text. We use deterministic greedy decoding with a single returned sequence and parse the JSON entry associated with each person. Invalid or missing entries are counted as failed predictions and assigned the image center $[0.5, 0.5]$ as the prediction. Due to API cost constraints, Gemini-3.6-Flash is evaluated on a randomly sampled subset of 1.5K test instances.

As shown in Table~\ref{tab:supp_zeroshot_vlm}, localization generally improves with model scale, yet even the strongest models remain far from task-specific performance: the best zero-shot result (0.231 Avg Dist. by Gemini-3.6-Flash) is more than twice the error of OmniGF (0.091). This indicates that, while large foundation models exhibit some zero-shot understanding of gaze, they still struggle with the precise continuous spatial localization required for gaze following, motivating the dual-branch design of OmniGF.

\begin{table}[h]
\centering
\small
\caption{Zero-shot gaze target estimation of VLMs on the GazeFollow test set. $^\dagger$ Evaluated on a randomly sampled subset of 1.5K test instances due to API cost constraints. OmniGF is included for reference.}
\label{tab:supp_zeroshot_vlm}
\begin{tabular}{lcc}
\toprule
Model & Avg Dist.$\downarrow$ & Min Dist.$\downarrow$ \\
\midrule
Qwen3-VL-2B-Instruct \cite{bai2025qwen3}  & 0.341 & 0.255 \\
Qwen3-VL-4B-Instruct \cite{bai2025qwen3}  & 0.340 & 0.258 \\
Qwen3-VL-8B-Instruct \cite{bai2025qwen3}  & 0.286 & 0.202 \\
Qwen3-VL-32B-Instruct \cite{bai2025qwen3} & 0.252 & 0.174 \\
InternVL3.5-8B \cite{wang2025internvl3_5}        & 0.271 & 0.202 \\
Gemini-3.6-Flash$^\dagger$ \cite{gemini36flash} & 0.231 & 0.165 \\
\midrule
OmniGF (Qwen3-VL-4B, fine-tuned) & 0.091 & 0.040 \\
\bottomrule
\end{tabular}
\end{table}

\section{Ablation on Loss Coefficients}
\label{sec:appendix_loss_coef}

We study the sensitivity of OmniGF to the coefficients of the two primary loss terms in the training objective (Sec.~\ref{sec:training_objectives}): the language modeling loss weight $\lambda_{lm}$ and the heatmap loss weight $\lambda_{hm}$. We vary one coefficient at a time while keeping all other settings fixed, and train on GazeFollow.

As shown in Table~\ref{tab:supp_loss_coef}, the model is robust to moderate changes of $\lambda_{lm}$, with Avg Dist. varying between 0.091 and 0.095. The heatmap loss weight has a larger effect: a small $\lambda_{hm}$ (0.1) noticeably degrades localization (0.102 Avg Dist.), as the dense spatial supervision is underweighted relative to the language modeling objective, while an overly large weight (100) also slightly hurts performance. Our default setting ($\lambda_{lm}=1.0$, $\lambda_{hm}=10.0$) achieves the best Avg Dist. and Min Dist. Variations in AUC across settings are small (within 0.008).

\begin{table}[h]
\centering
\small
\caption{Ablation of loss coefficients on GazeFollow. The default setting used in all other experiments is shaded. Best numbers are marked as bold.}
\label{tab:supp_loss_coef}
\begin{tabular}{ccccc}
\toprule
$\lambda_{lm}$ & $\lambda_{hm}$ & Avg Dist.$\downarrow$ & Min Dist.$\downarrow$ & AUC$\uparrow$ \\
\midrule
0.5 & 10.0  & 0.095 & 0.041 & 0.943 \\
\rowcolor{gray!15} 1.0 & 10.0  & \textbf{0.091} & \textbf{0.040} & 0.941 \\
2.0 & 10.0  & 0.092 & \textbf{0.040} & 0.942 \\
\midrule
1.0 & 0.1   & 0.102 & 0.047 & 0.936 \\
1.0 & 1.0   & 0.094 & 0.041 & \textbf{0.944} \\
\rowcolor{gray!15} 1.0 & 10.0  & \textbf{0.091} & \textbf{0.040} & 0.941 \\
1.0 & 100.0 & 0.096 & 0.042 & 0.942 \\
\bottomrule
\end{tabular}
\end{table}

\section{Scalability and Runtime Analysis}
\label{sec:appendix_runtime}

A key property of OmniGF is that all individuals in a scene are processed jointly, so the scene is encoded once per image regardless of the number of people, rather than once per person as in most prior gaze following methods. To quantify the resulting scaling, we measure the inference time and computational cost (TFLOPs) of OmniGF for images with varying numbers of people $N$. To obtain a controlled curve up to $N=8$, beyond the person counts available in most test frames, we attach $N$ synthetic head boxes to a real image and fix the generation length to 25 tokens per person, the length of one in-frame prediction. Runtime is reported per image, measured while processing batches of 64 images on a single NVIDIA RTX PRO 6000 Blackwell GPU with bfloat16 mixed precision.

As shown in Table~\ref{tab:supp_runtime}, when the number of people increases from 1 to 8, the computational cost increases by only $1.92\times$, rather than the $8\times$ incurred by running the backbone once per person, and the per-image runtime in this batched setting increases by $3.28\times$. Consequently, the per-person cost drops from 7.28 to 1.74 TFLOPs, and the per-person runtime from 53.62\,ms to 22.02\,ms. We note that the absolute inference cost of OmniGF remains higher than that of smaller task-specific models due to its 4B-parameter VLM backbone, although only 29.62M parameters are trainable through parameter-efficient fine-tuning. We discuss this limitation further in Appendix~\ref{sec:limitations}.

\begin{table}[h]
\centering
\small
\caption{Inference runtime and computational cost of OmniGF with varying numbers of people $N$ in an image, using synthetic head boxes and 25 generated tokens per person. Runtime is per image, measured with batches of 64 images. Per-person values are the totals divided by $N$.}
\label{tab:supp_runtime}
\begin{tabular}{lcccc}
\toprule
 & $N=1$ & $N=2$ & $N=4$ & $N=8$ \\
\midrule
Runtime (ms)            & 53.62 & 71.14 & 102.59 & 176.14 \\
Per-person runtime (ms) & 53.62 & 35.57 & 25.65  & 22.02 \\
\midrule
TFLOPs                  & 7.28  & 8.22  & 10.10  & 13.95 \\
Per-person TFLOPs       & 7.28  & 4.11  & 2.53   & 1.74 \\
\bottomrule
\end{tabular}
\end{table}

\section{Qualitative Results on Auxiliary Tasks} \label{sec:qualitative_supp}

In this section, we present qualitative visualizations across two extended gaze-following tasks: semantic gaze prediction and social gaze prediction.

\begin{figure}[t]
\setlength\tabcolsep{3pt}
\centering
\begin{tabular}{cccc}
 \includegraphics[width=0.24\linewidth, height=1.0in]{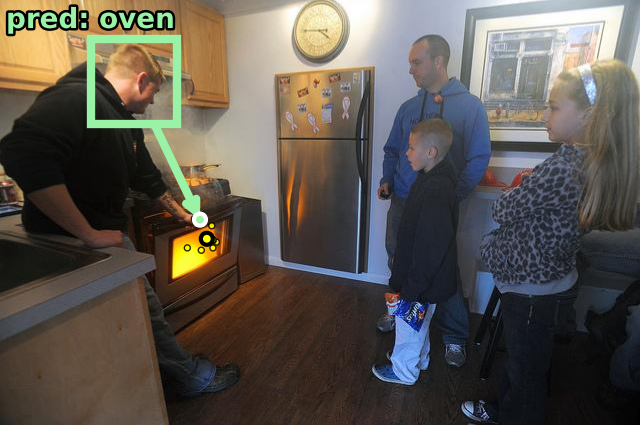} &
 \includegraphics[width=0.24\linewidth, height=1.0in]{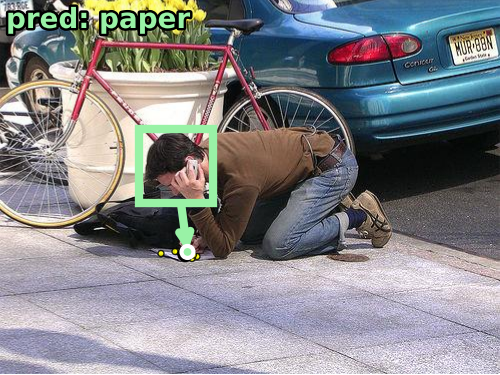} &
 \includegraphics[width=0.24\linewidth, height=1.0in]{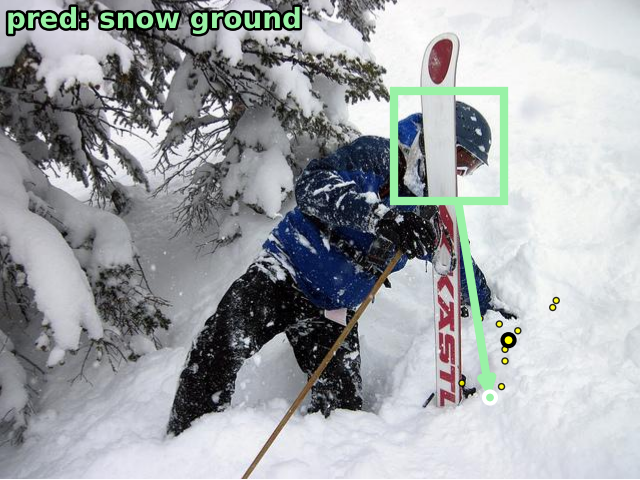} &
 \includegraphics[width=0.24\linewidth, height=1.0in]{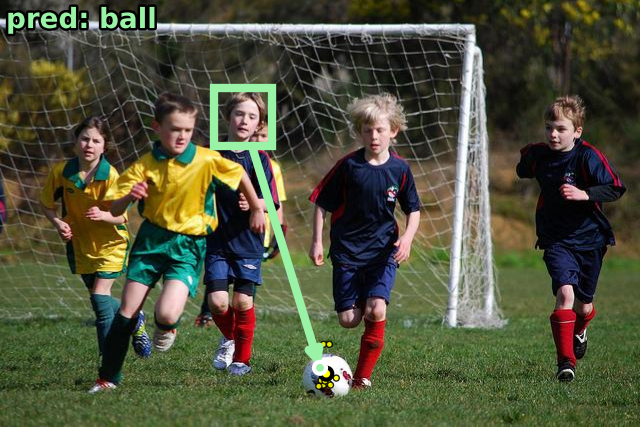} \\
 \includegraphics[width=0.24\linewidth, height=1.8in]{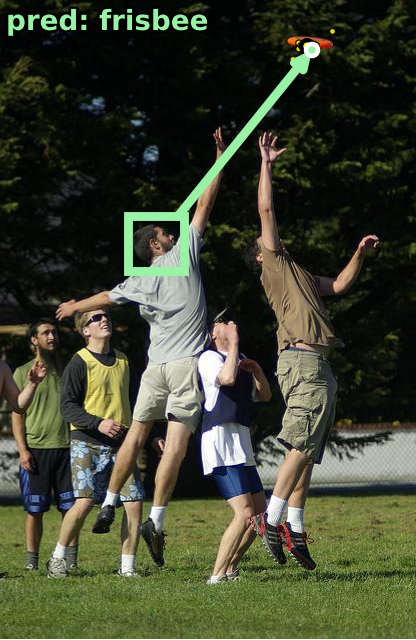} &
 \includegraphics[width=0.24\linewidth, height=1.8in]{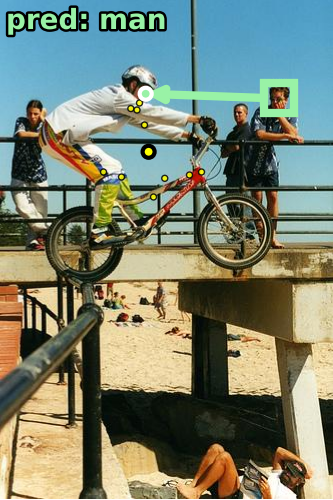} &
 \includegraphics[width=0.24\linewidth, height=1.8in]{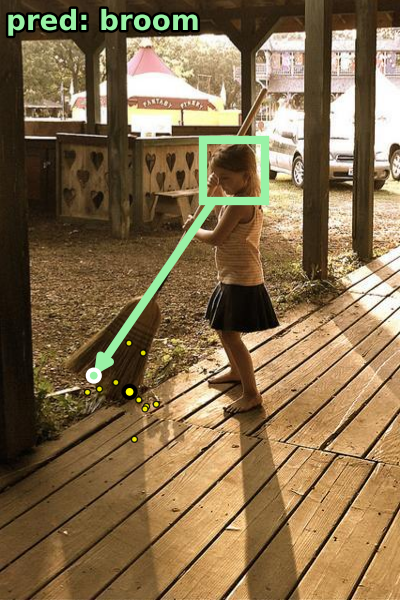} &
 \includegraphics[width=0.24\linewidth, height=1.8in]{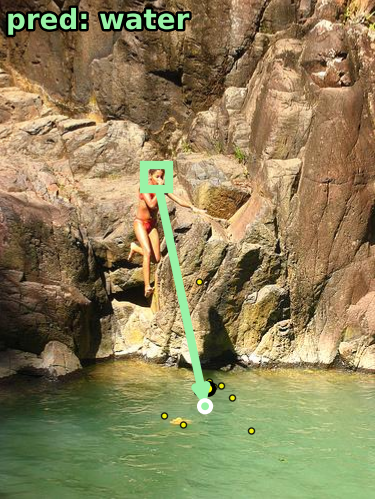} \\
  \includegraphics[width=0.24\linewidth, height=1.0in]{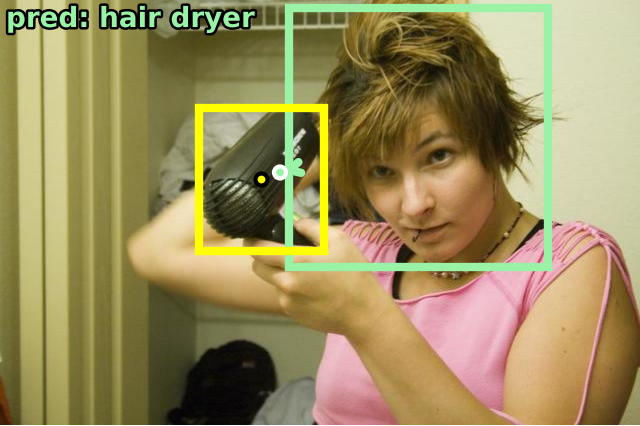} &
 \includegraphics[width=0.24\linewidth, height=1.0in]{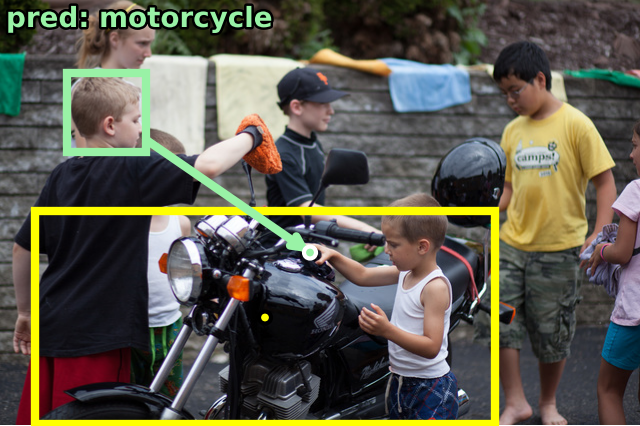} &
 \includegraphics[width=0.24\linewidth, height=1.0in]{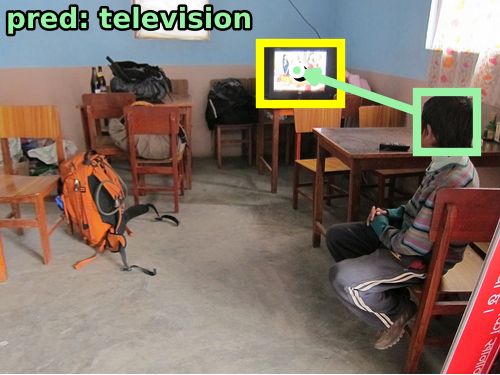} &
 \includegraphics[width=0.24\linewidth, height=1.0in]{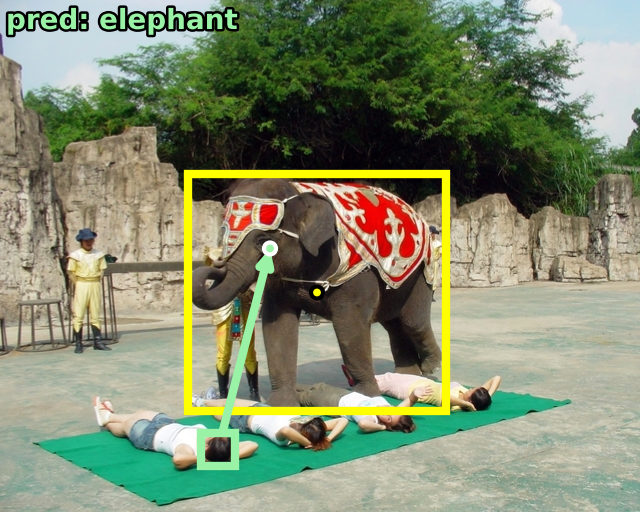} \\
 \includegraphics[width=0.24\linewidth, height=1.5in]{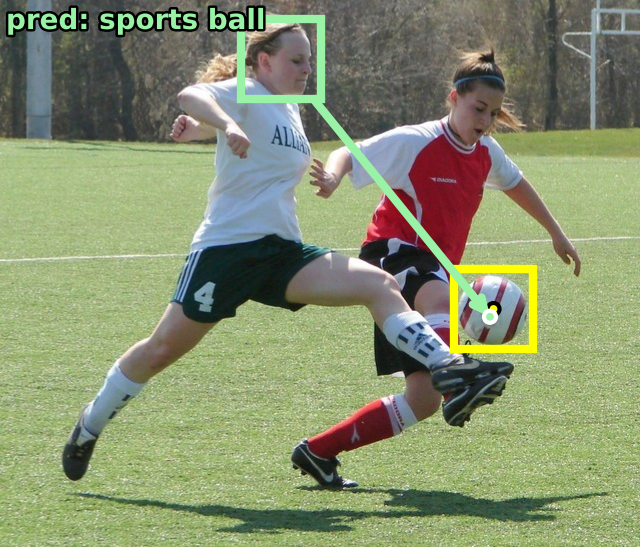} &
 \includegraphics[width=0.24\linewidth, height=1.5in]{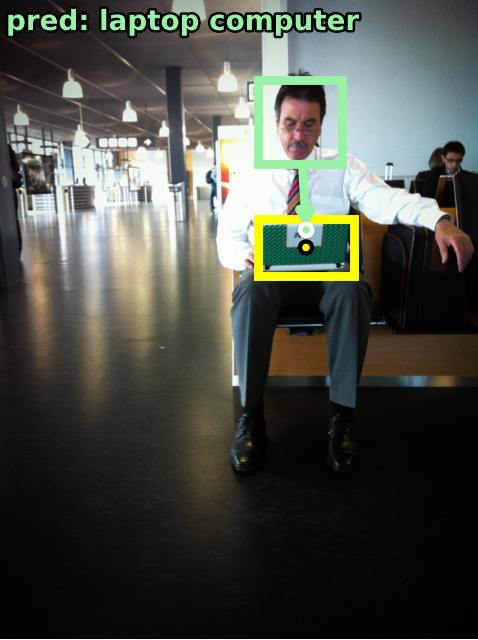} &
 \includegraphics[width=0.24\linewidth, height=1.5in]{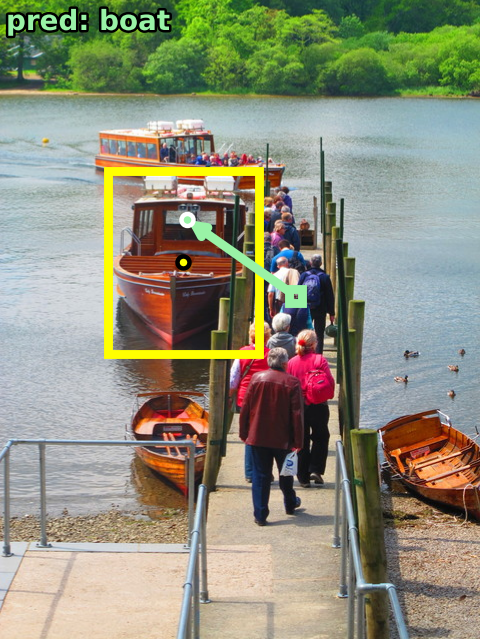} &
 \includegraphics[width=0.24\linewidth, height=1.5in]{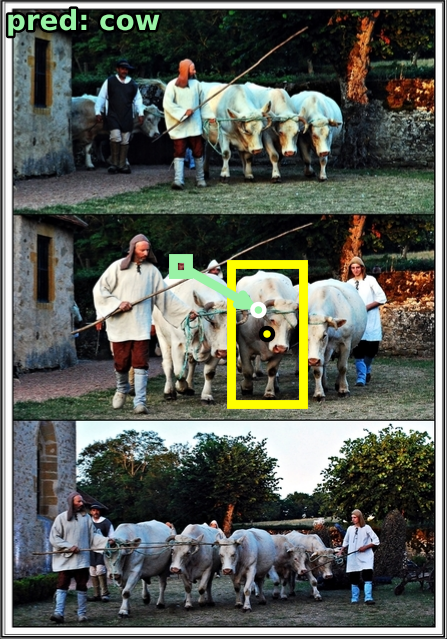} \\
\end{tabular}
\caption{Visualizations of the model predictions with semantic gaze labels on GazeFollow (first two rows) and GazeHOI (last two rows). Yellow dots or bounding boxes indicate the ground truth gaze target. Note that the GazeHOI dataset marks the ground truth gaze target coordinate as the bounding box center of the gaze target object.}
\label{fig:qualitative_vis_sem}
\end{figure}

In Fig. \ref{fig:qualitative_vis_sem}, we visualize the predicted gaze targets alongside their predicted semantic labels on the GazeFollow and GazeHOI datasets. As demonstrated, our model accurately localizes the spatial coordinate of the gaze target while simultaneously predicting highly context-aware semantic labels.

\begin{figure}[t]
\setlength\tabcolsep{3pt}
\centering
\begin{tabular}{cccc}
 \includegraphics[width=0.32\linewidth]{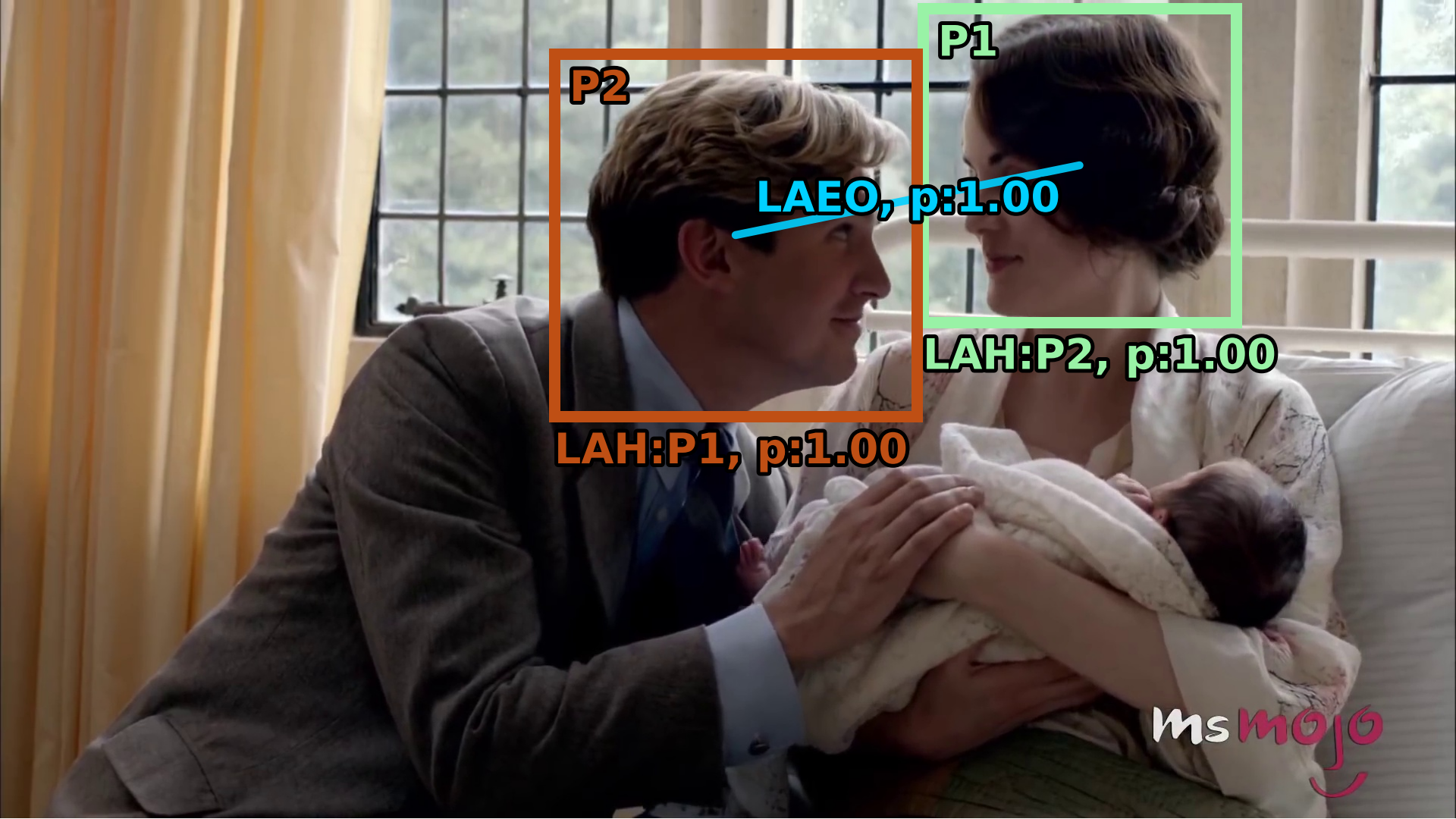} &
 \includegraphics[width=0.32\linewidth]{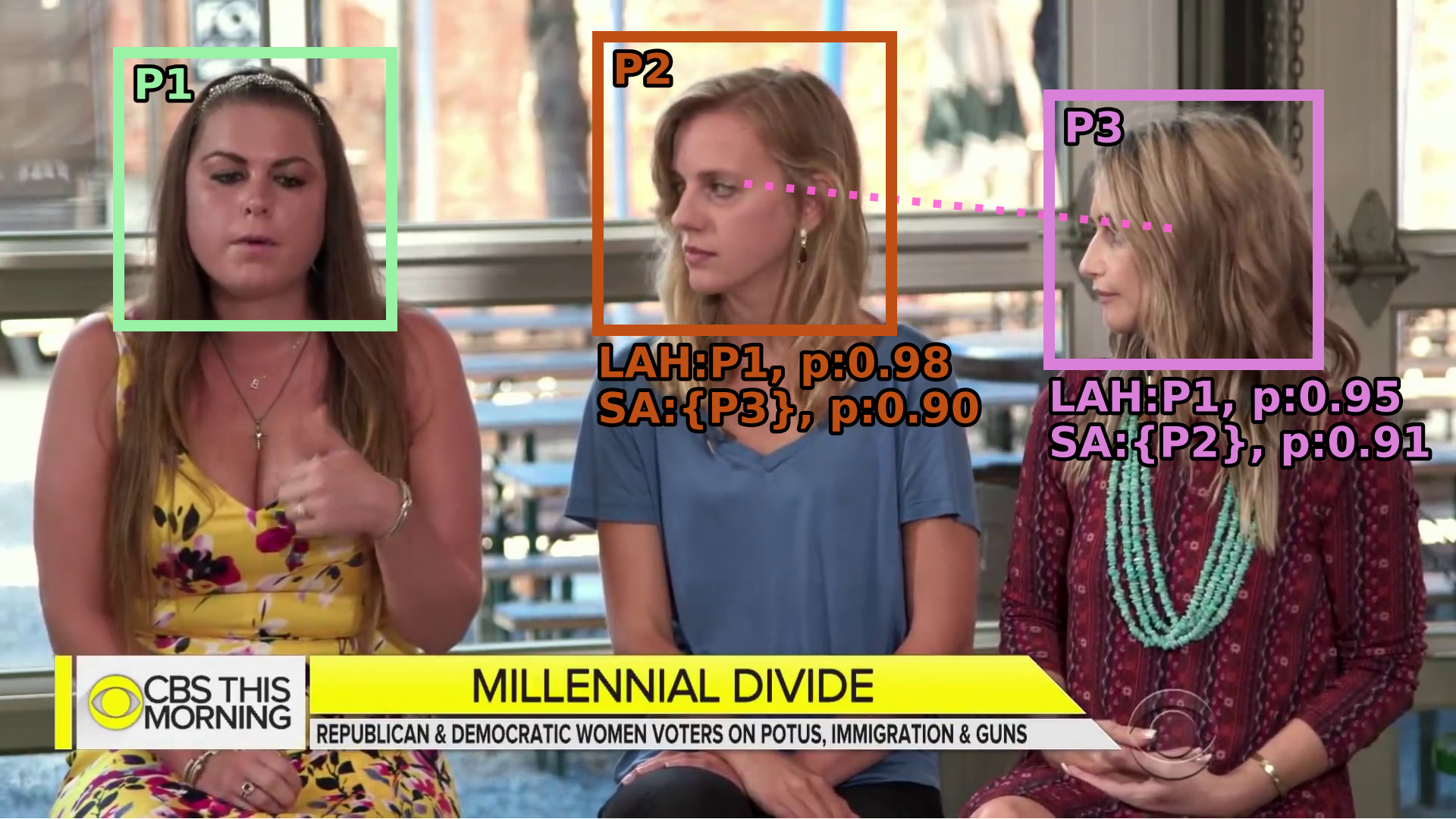} &
 \includegraphics[width=0.32\linewidth]{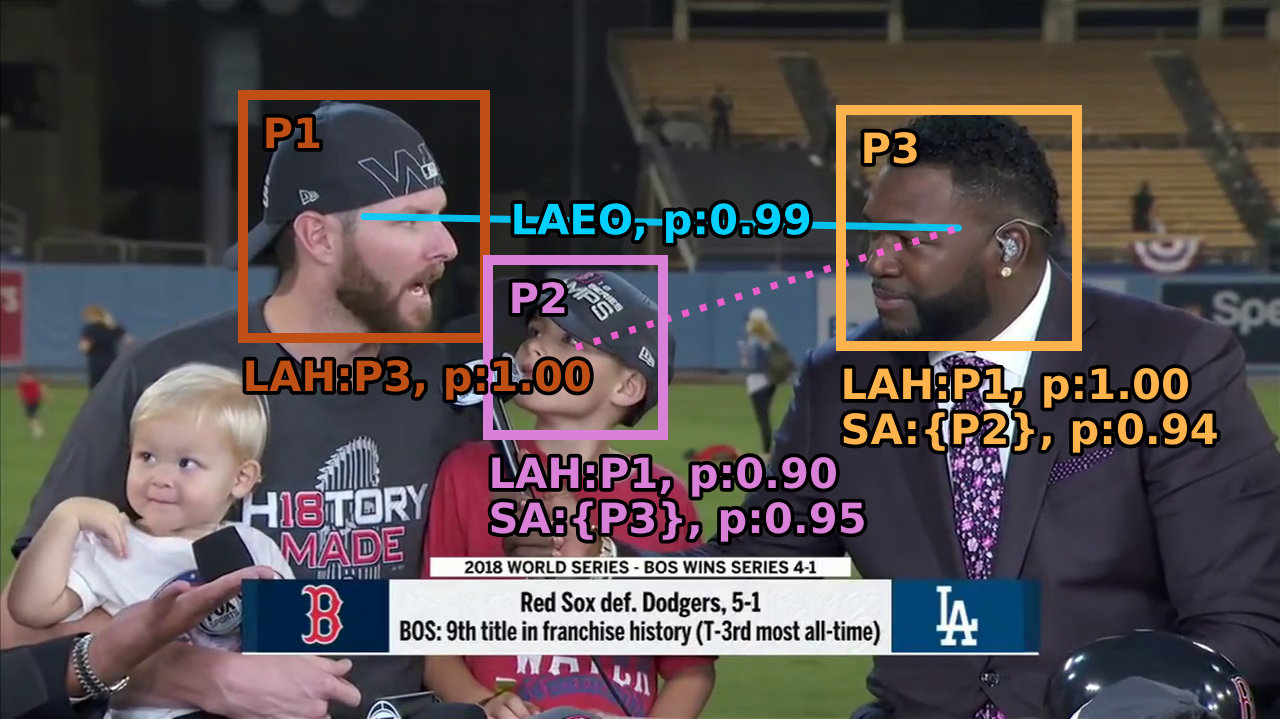} &\\
  \includegraphics[width=0.32\linewidth]{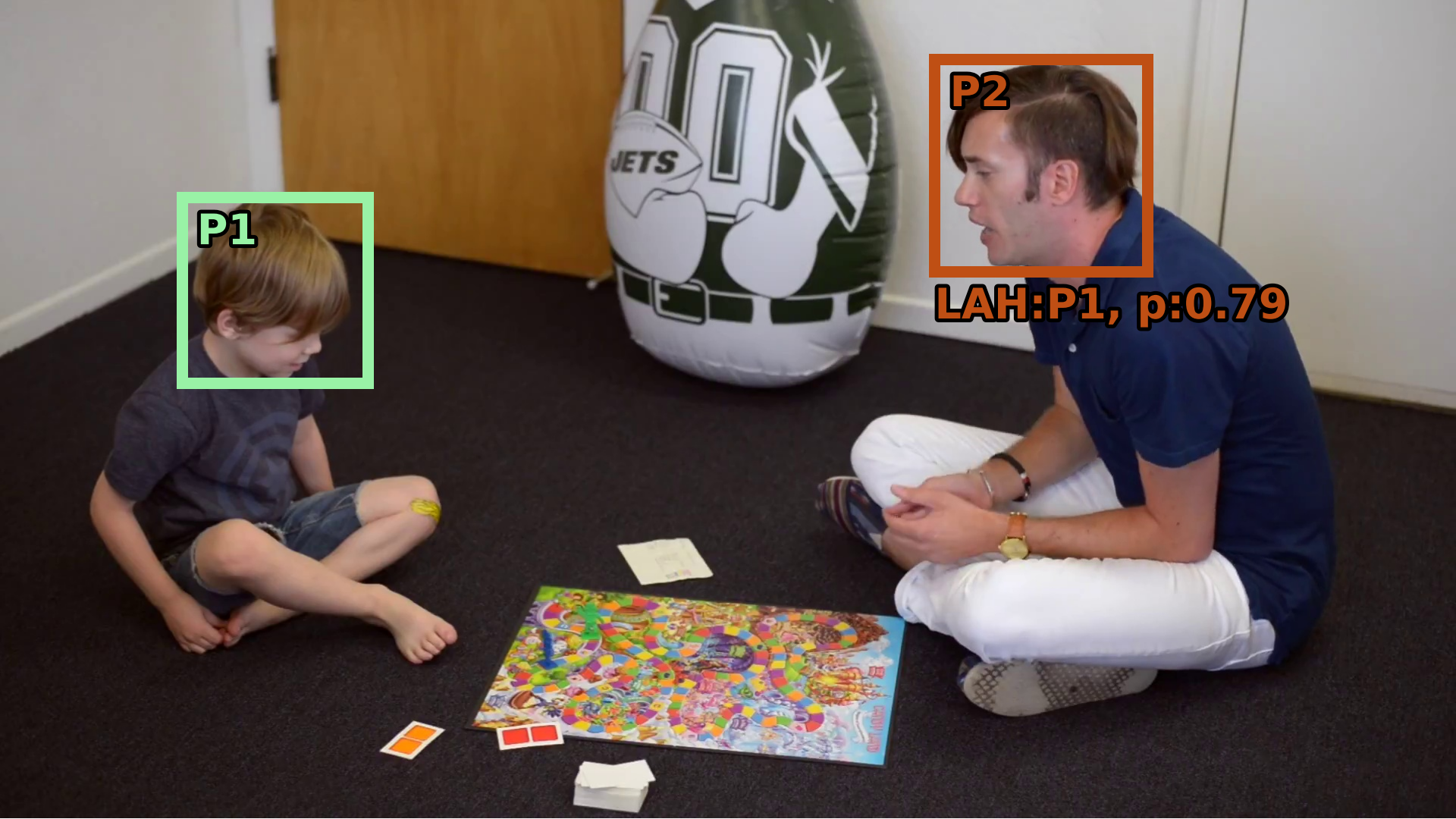} &
 \includegraphics[width=0.32\linewidth]{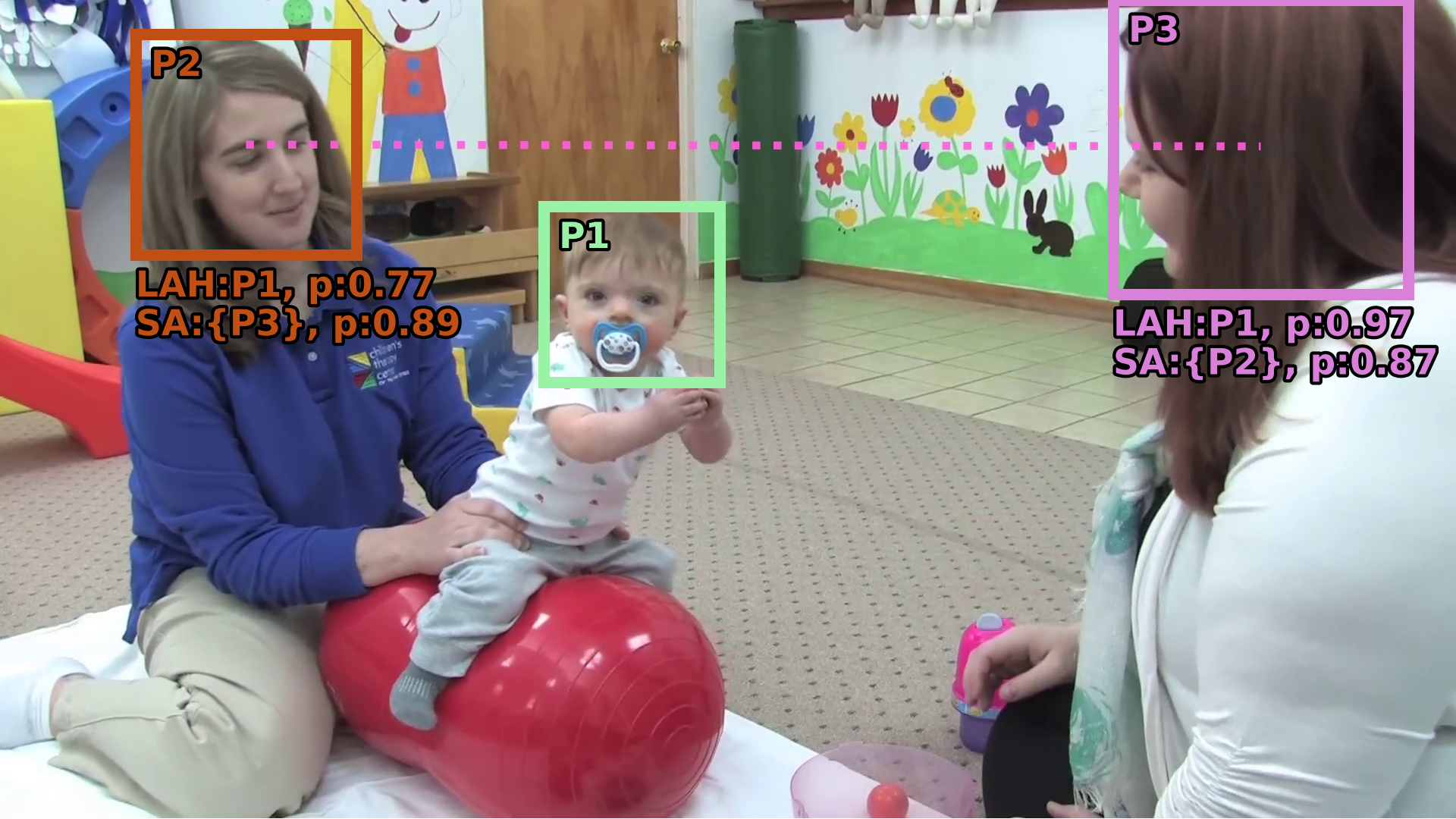} &
 \includegraphics[width=0.32\linewidth]{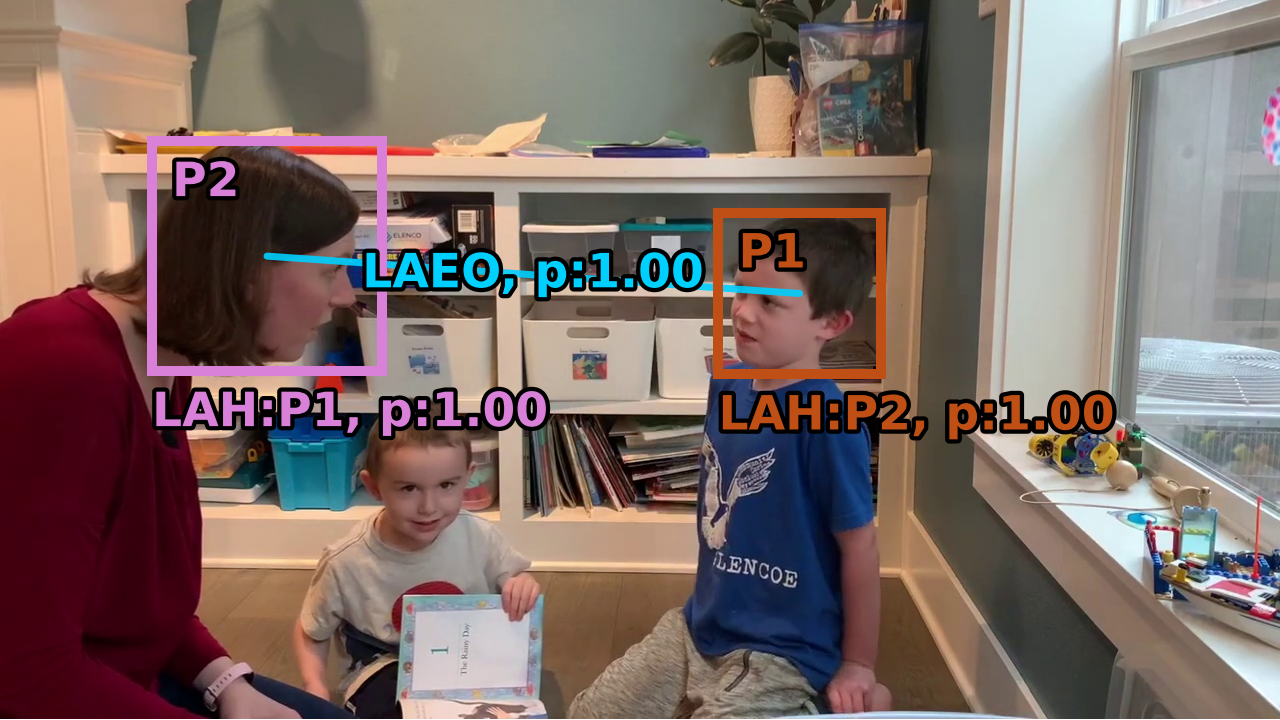} &\\
 \includegraphics[width=0.32\linewidth]{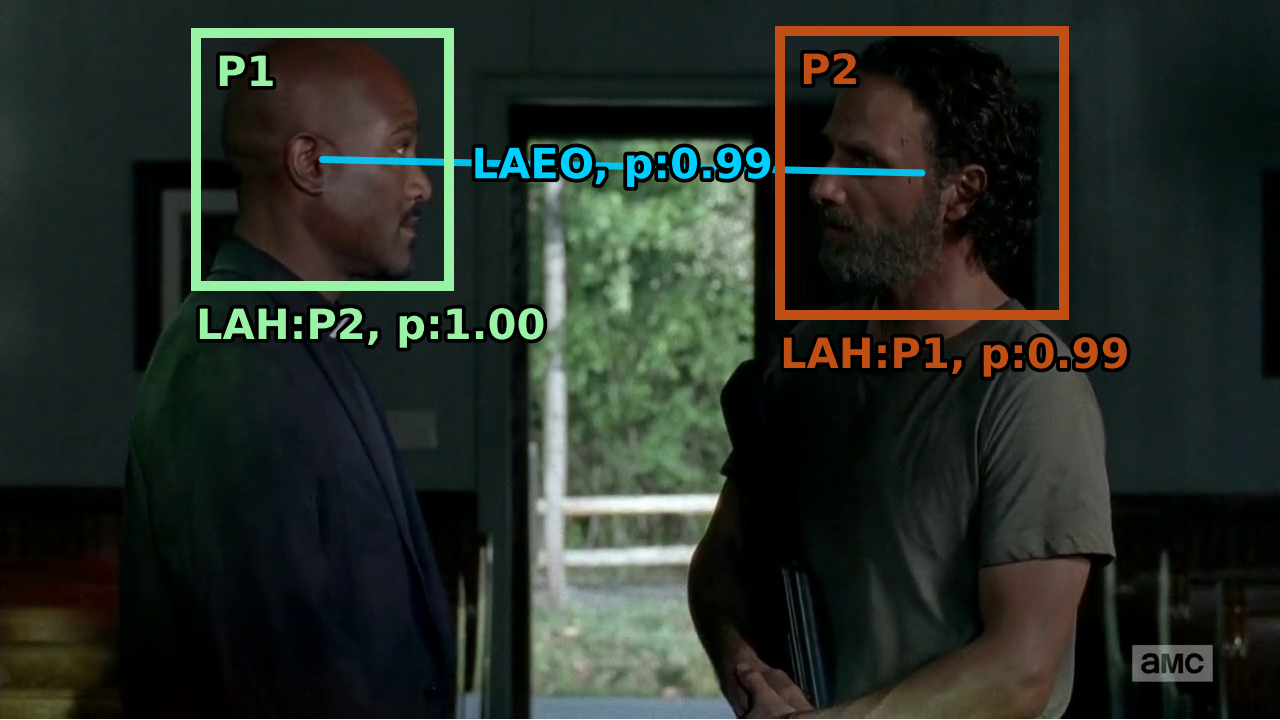} &
 \includegraphics[width=0.32\linewidth]{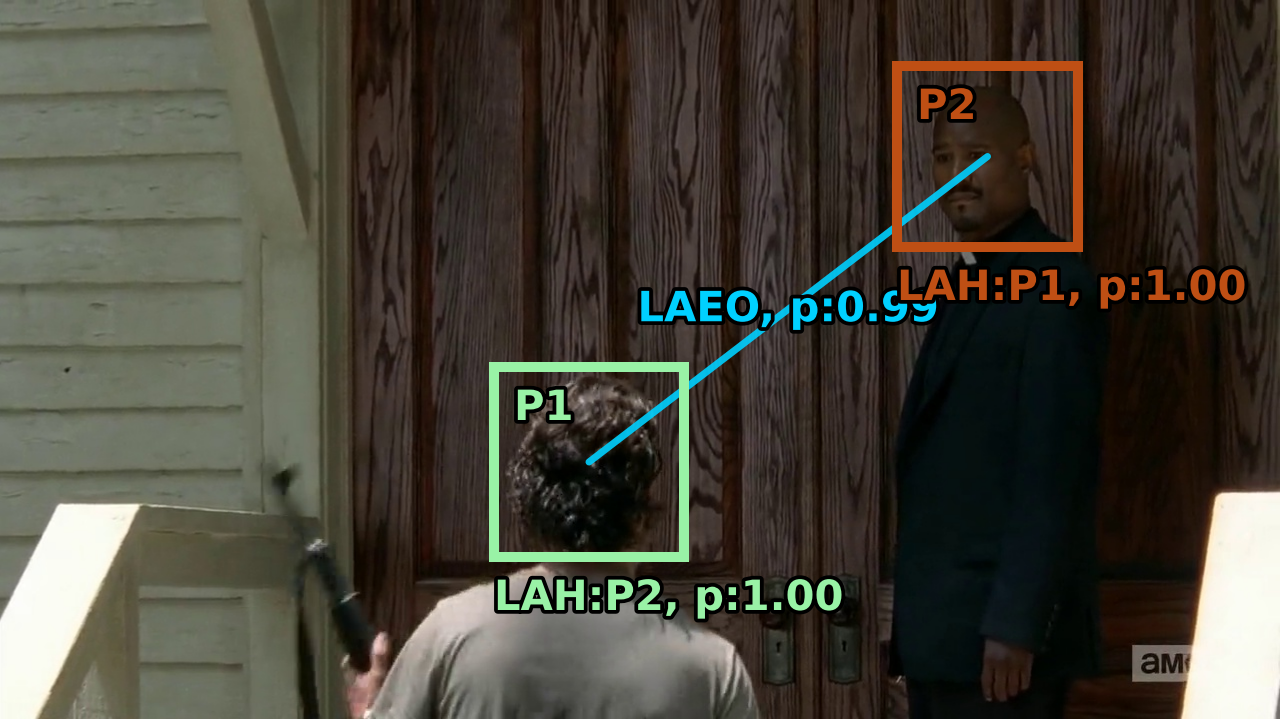} &
 \includegraphics[width=0.32\linewidth]{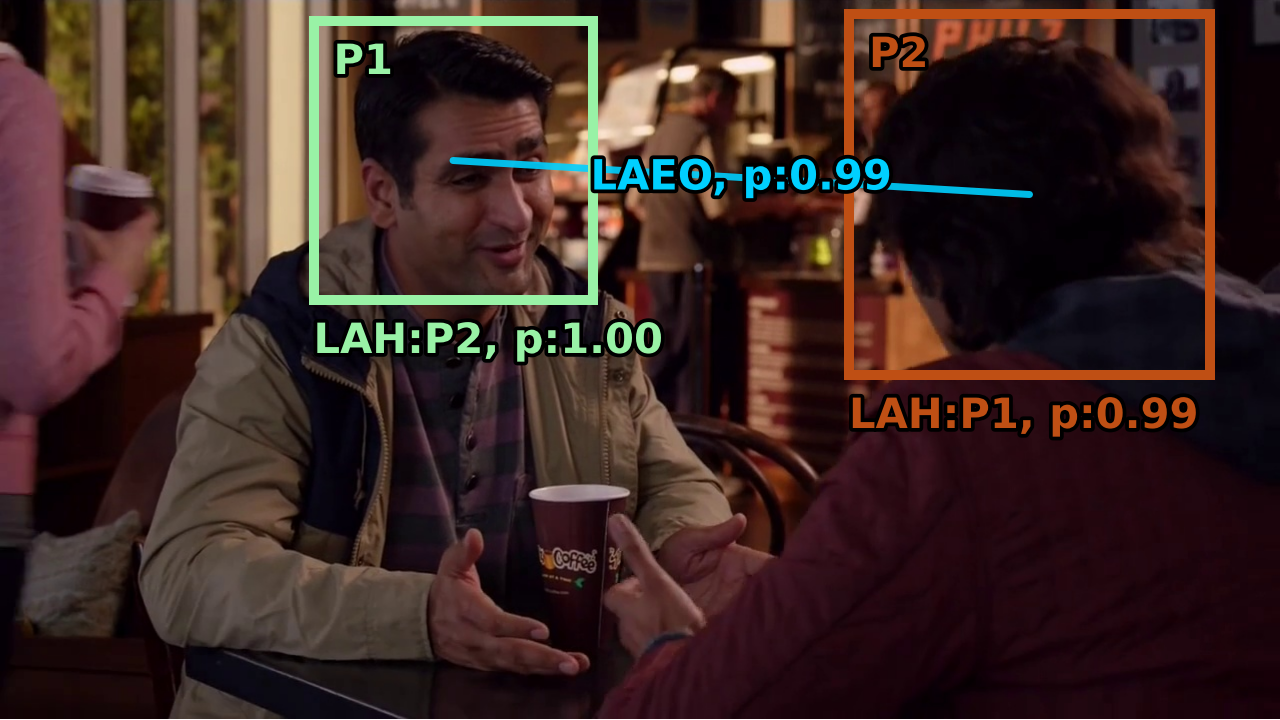} &\\
 \includegraphics[width=0.32\linewidth]{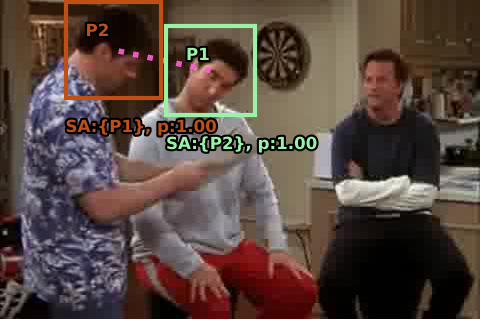} &
 \includegraphics[width=0.32\linewidth]{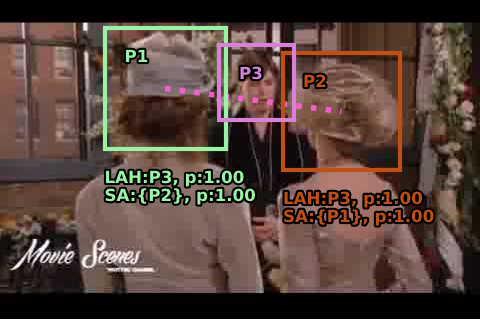} &
 \includegraphics[width=0.32\linewidth]{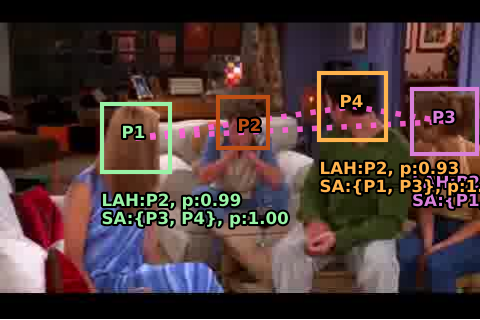} &\\
\end{tabular}
\caption{Qualitative results of social gaze predictions on the VSGaze dataset. Each row presents examples from a constituent sub-dataset: VideoAttentionTarget (Row 1), ChildPlay (Row 2), UCO-LAEO (Row 3), and VideoCoAtt (Row 4). Pairs engaging in Looking At Each Other (LAEO) are connected by solid blue lines, while individuals exhibiting Shared Attention (SA) are connected by dashed purple lines.}
\label{fig:qualitative_vis_social}
\end{figure}

In Fig. \ref{fig:qualitative_vis_social}, we illustrate the predicted social gaze labels across diverse interaction scenarios in the VSGaze dataset. The visualizations demonstrate the model's capacity to robustly predict multi-person social gaze dynamics and complex inter-subject gaze relationships across various challenging sub-datasets.

\section{Limitations and Future Work}
\label{sec:limitations}

While OmniGF establishes state-of-the-art performance across multiple benchmarks, we acknowledge a few limitations that provide avenues for future research. First, although our framework achieves single-pass multi-person inference, its reliance on a foundational VLM backbone inherently demands a larger memory footprint and computational budget compared to legacy CNN-based architectures, and its autoregressive decoding time grows with the number of people in the scene. Future deployments on edge devices could benefit from model quantization or distillation. Second, OmniGF is fundamentally an image-based architecture. When evaluated on video datasets (such as VideoAttentionTarget and ChildPlay), it processes scenes strictly on a frame-by-frame basis without explicit temporal modeling. While it cannot currently leverage motion cues or temporal smoothing to resolve highly ambiguous dynamic gaze behaviors, it is noteworthy that our frame-level approach still outperforms prior methods that explicitly incorporate temporal fusion. Future extensions could integrate temporal information to further boost robustness in video domains. Finally, while our dynamic gaze token injection scales elegantly to multi-person scenes, localization performance can experience minor drops in extremely crowded environments (e.g., scenes with over 15 individuals in VSGaze). In such extreme cases, the extended input token sequence may lead to attention dilution. Nevertheless, because the scene is encoded only once per image, the computational cost of OmniGF grows much more slowly with the number of people than the person-by-person processing required by prior multi-branch architectures (Appendix~\ref{sec:appendix_runtime}).

\section{Broader Impacts}
\label{sec:broader_impacts}

Our proposed framework, OmniGF, advances the capabilities of Vision-Language Models in understanding human gaze and complex social interactions. On the positive side, high-precision gaze estimation holds significant potential for beneficial applications, such as developing more intuitive human-computer interfaces, enabling socially aware collaborative robotics, and providing assistive technologies for individuals with disabilities. Furthermore, it can serve as a valuable analytical tool for behavioral research in psychology and healthcare, potentially aiding in the study of neurodevelopmental conditions such as autism. However, we also acknowledge the potential negative societal impacts of this technology. The ability to accurately track and analyze where individuals are looking in dense, multi-person scenes raises privacy concerns. If deployed irresponsibly, such systems could be misused for intrusive public surveillance, non-consensual tracking of consumer attention in retail environments, or pervasive employee monitoring. Additionally, because our method relies on foundational VLMs, it may inherit or amplify underlying biases present in the large-scale pre-training data, which could lead to disparate performance across different demographic groups. We encourage the community to establish strict ethical guidelines and prioritize privacy-preserving mechanisms when deploying gaze-aware systems in real-world environments.


\end{document}